%% file: main.tex
\documentclass[aps,prx,reprint,superscriptaddress,amsmath,amssymb,longbibliography,nofootinbib,floatfix]{revtex4-2}
\usepackage{graphicx}
\usepackage{xcolor}
\usepackage{multirow}
\usepackage[colorlinks=true,linkcolor=blue,citecolor=blue,urlcolor=blue]{hyperref}

\newcommand{\pt}{$p_\mathrm{T}$}

\begin{document}

\title{Contrastive Learning for Interpretable Anomaly Detection at Collider Experiments}

\author{Haoyi Jia}
\email{hjia625@slac.stanford.edu}
\affiliation{Department of Physics, Stanford University, 450 Jane Stanford Way, Stanford, CA 94305}
\affiliation{SLAC National Accelerator Laboratory, 2575 Sand Hill Rd, Menlo Park, CA 94025}

\author{Sagar Addepalli}
\affiliation{SLAC National Accelerator Laboratory, 2575 Sand Hill Rd, Menlo Park, CA 94025}

\author{Julia Gonski}
\affiliation{SLAC National Accelerator Laboratory, 2575 Sand Hill Rd, Menlo Park, CA 94025}

\date{\today}

\begin{abstract}
Generic event-level anomaly detection for collider physics has two recurring problems: anomaly scores are hard to interpret, and they correlate strongly with energy scale and object multiplicity.
We present \textbf{O}rganized \textbf{R}epresentation via \textbf{C}ontrastive learning for \textbf{A}nomaly detection (\textbf{ORCA}), a two-stage framework that first learns an embedding space via supervised contrastive learning across a diverse set of physics processes, then runs a standard autoencoder in that space to generate event-level anomaly scores.
On a simulated dataset consistent with conditions at the High-Luminosity Large Hadron Collider, ORCA delivers significant gains in both breadth and depth of sensitivity to new physics signals with respect to a baseline autoencoder architecture.
Beyond improved sensitivity, the contrastive embedding makes the anomalous sample interpretable: because known processes occupy distinct regions of the space, a maximum-likelihood template fit to the embedding distributions can attribute events in an anomalous sample to template physics processes with quantified uncertainties.
We demonstrate that the fit accurately recovers injected signal yields, including for signals excluded from the training of the embedding, and characterizes signals absent from the template library through the known processes they most resemble.
These results establish ORCA as a route to interpretable anomaly detection-based searches at colliders, where the embedding geometry carries higher dimensional physics information compared to standard one-dimensional output fits, enhancing downstream statistical analysis.
\end{abstract}

\maketitle

\input{sections/intro}
\input{sections/methods}
\input{sections/results}
\input{sections/conclusions}

\section*{Declaration of generative AI and AI-assisted technologies}

During the preparation of this work, the authors used Claude and Claude Code (Anthropic), based on the \texttt{Claude Opus 4.8} and \texttt{Claude Fable 5} language models, to assist with writing analysis code and with language editing. After using these tools, the authors reviewed and edited the content as needed and take full responsibility for the content of the published article.

\begin{acknowledgments}
The authors thank Peter Graham for helpful discussions on the theoretical interpretation of these results. This work was supported by the U.S. Department of Energy under contract number DE-AC02-76SF00515.
\end{acknowledgments}

% Bibliography: REVTeX 4.2 automatically selects the apsrev4-2 style
\bibliography{biblio}

\end{document}

%% file: sections/intro.tex
\section{Introduction}
\label{sec:intro}

%- P1: Why/what is AD at HEP? 
Anomaly detection (AD) has been established as a key machine learning (ML) technique in searching for signs of beyond the Standard Model (BSM) physics at the Large Hadron Collider (LHC)~\cite{Belis_2024,Kasieczka_2021,Aarrestad_2022}. AD methods differ from traditional analysis techniques in their ability to be signal model-agnostic while retaining high background suppression power, allowing for a large catchment area of BSM phenomena that could enter the measurement phase space. 
The ATLAS~\cite{HDBS-2018-59,HDBS-2019-23,EXOT-2022-07,EXOT-2021-34} and CMS~\cite{cmsad} experiments have employed promising deep learning-based AD techniques in searches for exotic phenomena across a range of final states.

AD subsumes a large class of ML techniques and models, each with different strengths and limitations~\cite{amram2026modelagnosticsignaldiscoverymachine}. 
While all aim for a degree of model independence not afforded by traditional supervised ML methods, anomaly detection tools vary in their use of signal models. 
Unsupervised techniques train using unlabeled data only, and do not make any assumptions about the nature of the signal~\cite{Heimel_2019,Cheng_2023,Kahn_2021,Farina_2020}. 
Self-supervised techniques also primarily train using unlabeled data~\cite{SciPostPhysCore.7.3.056, SciPostPhys.18.2.042}, but construct pseudo-labels from the data itself to shape an intermediate representation. 
A third class of partially supervised techniques assists this learning with labeled data~\cite{li2026signalawarecontrastivelatentspaces,5n77-ynsp, elsharkawy2025contrastivenormalizingflowsuncertaintyaware}, using a supervised representation-learning stage that precedes an unsupervised AD stage from which the labels are withheld.

%- P2: limitations of AD: 
While AD methods have been shown to be powerful in isolating signal-rich phase spaces without explicitly relying on the signal modeling itself, there are limitations in their traditional implementations. 
The first kind of limitation is learning capacity. 
Compression-based techniques, such as bare autoencoders~\cite{6302929}, when trained directly on low-level inputs, are sensitive to training population characteristics. 
A reconstruction-based training objective incentivizes the model to memorize recurring entries, driving the net loss down faster than learning higher-order correlations would. 
Raw input values, dataset topology, and neural-network interpolation biases can all cause normal events to reconstruct poorly and rare true anomalies to reconstruct well~\cite{Batson_2021}. 
Such limitations also add dependence on detector conditions such as pileup.

The second limitation of current AD methods concerns interpretation. Understanding the anomaly score behavior is typically limited to examining clustering and correlations relative to the raw kinematic inputs. Potentially more sophisticated interpretations can be made using Monte Carlo (MC) simulations, at the risk of such interpretations arising from mismodelings in MC rather than from genuine deviations.
Signal extraction in Refs.~\cite{HDBS-2018-59,HDBS-2019-23,EXOT-2022-07,EXOT-2021-34,cmsad} is performed either in an anomalous phase space where an experimentally measured quantity, such as an invariant mass distribution, is fit, or in a kinematically tight phase space where the anomaly score distribution is fit. 
Both of these approaches require specifying a relatively narrow measurement phase space, reducing the potential generality of AD. 
Furthermore, this lack of interpretability significantly reduces the value of AD results for model testing and development; with a single anomaly score, many distinct models, each spanning a broad range of free parameters, can manifest as indistinguishable score distributions.

%- P3: What is CL? 
Contrastive learning (CL)~\cite{3495724.3496555,3524938.3525087,chen2020bigselfsupervisedmodelsstrong} can address both of these limitations of traditional AD methods. CL refers to a family of representation-learning techniques in which an encoder is trained to map pairs of events designated as similar (positive pairs) close together in a fixed-dimensional vector space, while pushing apart pairs designated as dissimilar. The resulting representation is trained to encourage two properties: alignment and uniformity~\cite{3524938.3525859}, whereby positive pairs map close together while representations spread across the available space. Together these impose a context-specific structure beyond what the raw features provide. Positive pairs can be constructed either through truth-labeled data, referred to in literature as supervised CL~\cite{NEURIPS2020_d89a66c7}, or through artificial augmentations of individual events that leave the underlying physics invariant~\cite{3524938.3525087}, referred to as self-supervised CL. 

In AD applications for collider physics, this has been extended to augmentations chosen so that the physics is deliberately not invariant, which are instead used as repulsive pairs to sensitize the representation to generic anomalous features~\cite{SciPostPhysCore.7.3.056,SciPostPhys.18.2.042}. CL posits that using this vector space provides a richer alternative to using the raw input features for downstream AD. 
%- P4: Why CL for AD? 
Specifically, creating a well-constrained and controlled intermediate representation makes a density-estimation or reconstruction task much simpler. 
The statistics-driven artifacts that artificially bloat anomaly scores in models are dispersed in this representation, allowing for better learnability of subtle high-dimensional correlations, which are the primary target of AD-based searches.
Furthermore, a carefully constructed intermediate vector space can be used as a representation of the underlying physics, enabling a continuous handle for embedding (and hence extracting) physics model or process information. 

%- P5: This work/ORCA intro 
This work presents the \textbf{O}rganized  \textbf{R}epresentation via \textbf{C}ontrastive learning for \textbf{A}nomaly detection (\textbf{ORCA}) method. 
A first stage uses CL to create a physics process-based latent embedding, trained with supervision over background and signal processes using object-level kinematic features as inputs. 
This embedding is then used as an input to a second stage, namely an autoencoder trained without supervision on background events alone. 
The two stages are coupled only through the learned embedding, so each can be exchanged independently. The architectures adopted here represent one use case of the framework.
We hence show that by using a physics-informed embedding, we improve the discriminative power of the AD model as compared to a baseline autoencoder trained directly over the object-level features. 
Furthermore, we build a framework for signal extraction using the per-event embedding vector values, thus showing that CL can be leveraged for robust physics interpretation without losing phase space generality. 
This framework is structurally equivalent to the binned fitting techniques conventionally used in BSM searches. Interpreting the results therefore requires no additional tooling or familiarity with the underlying ML method, lowering the barrier to using published AD results from real data for reinterpretation.

%- P6: Related works in HEP.
Previous work on physics-informed embeddings for AD in high energy physics (HEP) has proceeded along two broad lines. The first imposes physical symmetries directly in the model architecture~\cite{Hao2023,SciPostPhys.16.3.062}. 
The second learns the embedding from data, as in self-supervised CL enforces approximate invariances via augmentations of jet or event constituents~\cite{SciPostPhys.12.6.188}, an approach adapted to AD through anomaly-motivated augmentations~\cite{SciPostPhysCore.7.3.056,SciPostPhys.18.2.042}. 
Refs.~\cite{5n77-ynsp,li2026signalawarecontrastivelatentspaces} construct contrastive embeddings from labeled physics processes and demonstrate substantial gains in detection sensitivity. Neither, however, develops the embedding into a framework for signal extraction or physics interpretation. 
A separate precedent for the two-stage structure of feature embedding for autoencoding can be found in Ref.~\cite{Matos_2025}, 
though its embedding objective uses classification rather than CL. 
This work presents a first application of a contrastive latent representation to the physics interpretation of anomalies, performed without narrowing to a dedicated measurement phase space.

%% file: sections/methods.tex
\section{Methodology}
\label{sec:methods}

\subsection{Samples}

ORCA is trained and evaluated on events from a simulated LHC collision dataset, COLLIDE-2V~\cite{Moreno2025COLLIDE2V}, comprising approximately 750 million proton-proton collisions at $\sqrt{s} = 13.6$ TeV.
The dataset contains over 50 distinct Standard Model processes generated with MadGraph~\cite{Alwall_2014} and Pythia~\cite{bierlich2022comprehensiveguidephysicsusage} under High-Luminosity LHC conditions, corresponding to an average number of simultaneous interactions $\mu$ equal to 200.
A simplified detector response is emulated using Delphes~\cite{delph2014} with a CMS Phase II configuration card that includes two event views; fully reconstructed objects assuming high-performance tracking, and custom trigger-level objects with degraded performance matching L1 trigger constraints.
In this work only fully reconstructed objects are considered.

Events are modeled by a 110-dimensional vector built from the 10 leading jets, four leading electrons, four leading muons, four leading photons, and the missing transverse energy (MET).
Objects of each type are ordered by \pt~and truncated to these fixed multiplicities; events with fewer objects are zero-padded, and any additional objects beyond the retained counts are discarded.
Jets are reconstructed using the anti-$k_t$ algorithm~\cite{MatteoCacciari_2008} with a radius of $R = 0.4$.
They are described by six features: transverse momentum \pt, pseudorapidity $\eta$, azimuthal angle $\phi$, b-tagging discriminant, electric charge, and mass.
Electron features include \pt, $\eta$, $\phi$, the ratio of hadronic to
electromagnetic calorimeter energy ($E_{\mathrm{had}}/E_\mathrm{em}$), and the $\rho$-corrected isolation variable; photons only use \pt, $\eta$, $\phi$.
Muon features include \pt, $\eta$, $\phi$, and the $\rho$-corrected isolation variable.
The missing transverse energy is encoded by its magnitude and azimuthal angle.

%-----------------------------------------------------------------------
\subsection{Machine Learning Models}

\begin{figure*}[!tb]
\centering
\includegraphics[width=\textwidth]{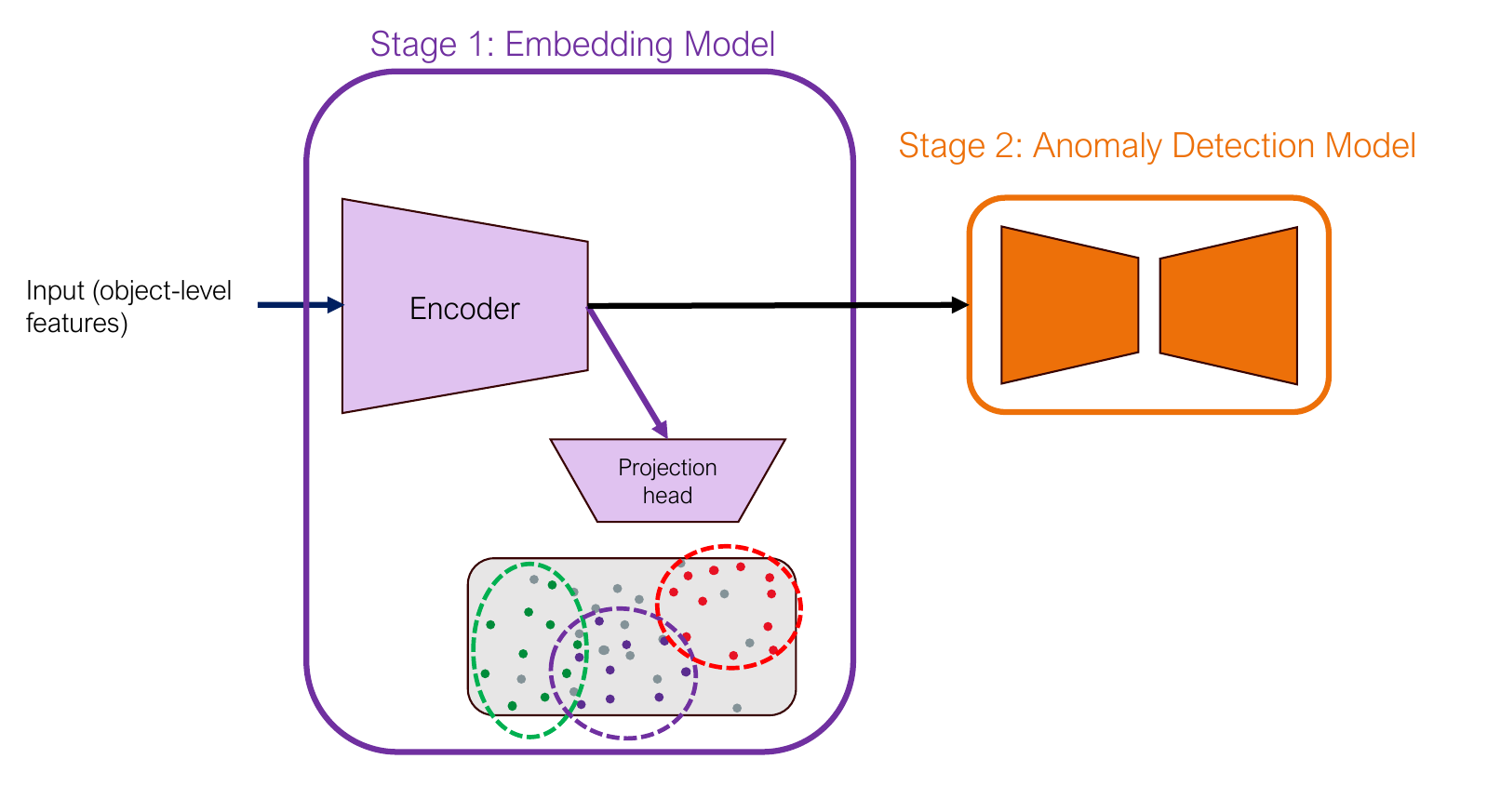}
\caption{Overview of the two-stage ORCA framework. In the first stage, an embedder consisting of an encoder and a projection head is trained with a supervised contrastive loss to produce a representation space in which the Standard Model process groups are separated from one another. In the second stage, an anomaly detection model, here an autoencoder, is trained unsupervised on background data embedded into this representation, and its reconstruction error serves as the anomaly score. The projection head is used only during the first-stage training.}
\label{fig:orca_overview}
\end{figure*}

The ORCA method is trained with a two stage process, illustrated in Figure~\ref{fig:orca_overview}.
First, an embedder model is trained to create a representation of input events informed by the use of a contrastive loss term over a variety of disparate signals.
Second, an unsupervised anomaly detection model is trained on data embedded into the representation as defined by the first stage.
A third model is developed and trained to provide context to the ORCA results, specifically an autoencoder trained directly over the physics inputs of the ORCA embedder model.
Details of the structure and training of each model are provided below.

%---------------
\paragraph{ORCA Stage 1: Embedder}

The embedder is composed of an encoder network and a projection network, inspired from Ref.~\cite{khosla2021supervisedcontrastivelearning}.
The encoder network maps each event $x$ to a $D_E$-dimensional representation vector $\vec{r}\in\mathcal{R}^{D_E}$.
Specifically, each object is first mapped into a common $D_{O}$-dimensional embedding space by a dedicated object-wise network, one per object type, consisting of stacked fully-connected layers with batch normalization and GELU activations, applied with weights shared across all objects of that type.
The resulting object tokens are augmented with a learnable positional embedding and processed by a stack of transformer encoder blocks, each comprising multi-head self-attention and a position-wise feed-forward network with GELU activations, residual connections, and layer normalization.
A learnable multi-head attention-pooling layer, in which a single trainable query attends over the encoded tokens, then aggregates the set into a single event representation $\vec{r}$, which is $\ell_2$-normalized onto the unit hypersphere.
A projection network maps the representation $\vec{r}$ to a lower $D_P$-dimensional vector $\vec{z}\in\mathcal{R}^{D_P}$ on which the contrastive loss is computed.
It is a multilayer perceptron with one hidden layer followed by a linear output layer, and its output is likewise $\ell_2$-normalized.
Following Ref.~\cite{khosla2021supervisedcontrastivelearning}, the projection network is used only during training; downstream anomaly detection operates on the event representation $\vec{r}$ rather than on $\vec{z}$.

The embedder model is trained with the following loss:
\begin{equation}
    \mathcal{L} = \mathcal{L}_{\mathrm{contra}} + \lambda\mathcal{L}_{\mathrm{var}},
\end{equation}
where:
\begin{itemize}
\item $\mathcal{L}_{\mathrm{contra}}$ is an InfoNCE~\cite{oord2019representationlearningcontrastivepredictive}-like supervised contrastive loss~\cite{khosla2021supervisedcontrastivelearning} that encourages separation of the physics groups in the projection space;
\begin{equation}
    \mathcal{L}_{\mathrm{contra}}
= -\sum_{i\in\mathcal{B}} \frac{1}{|P(i)|}\sum_{p\in P(i)}
\log\!
\frac{\exp\!\left(\vec{z_i} \cdot \vec{z_p} / \tau\right)}
{\sum\limits_{\substack{k\in \mathcal{B}\\k\neq i}} \exp\!\left(\vec{z_i} \cdot \vec{z_k} / \tau\right)},
\end{equation}
where $P(i)$ is the positive set for event $i$ containing all other events in the batch $\mathcal{B}$ belonging to the same process group; $\tau$ is a temperature hyperparameter that controls the sharpness of the similarity distribution, set to $\tau = 0.07$.
\item $\mathcal{L}_{\mathrm{var}}$ is a variance regularizer that encourages each dimension to maintain a target standard deviation and thereby prevents representational collapse, inspired by Ref.~\cite{bardes2022vicregvarianceinvariancecovarianceregularizationselfsupervised}.
This differs from a conventional Kullback-Leibler divergence loss in a variational autoencoder, which pushes the embedding dimensions towards a specific shape, namely a Gaussian prior.
Since the event representation is $\ell_2$-normalized, all points lie on the $D_E$-dimensional unit hypersphere.
The $1/\sqrt{d}$ term is the standard deviation per dimension of a uniform distribution of points on the $d$-dimensional unit hypersphere, with $d = D_E$.
A factor of $0.9$ is applied as a conservative margin, so that dimensions whose standard deviation exceeds this target do not contribute to the loss.
The hyperparameter $\lambda$ is optimized and ultimately set to 1.
\begin{equation}
    \mathcal{L}_{\mathrm{var}}=\frac{1}{d}\sum_{j=1}^{d}\left[\max\left(0,\,\frac{0.9}{\sqrt{d}}-\sigma_j\right)\right]^2
\end{equation}
\end{itemize}

\begin{figure*}[!tb]
\centering
\includegraphics[width=1.65\columnwidth]{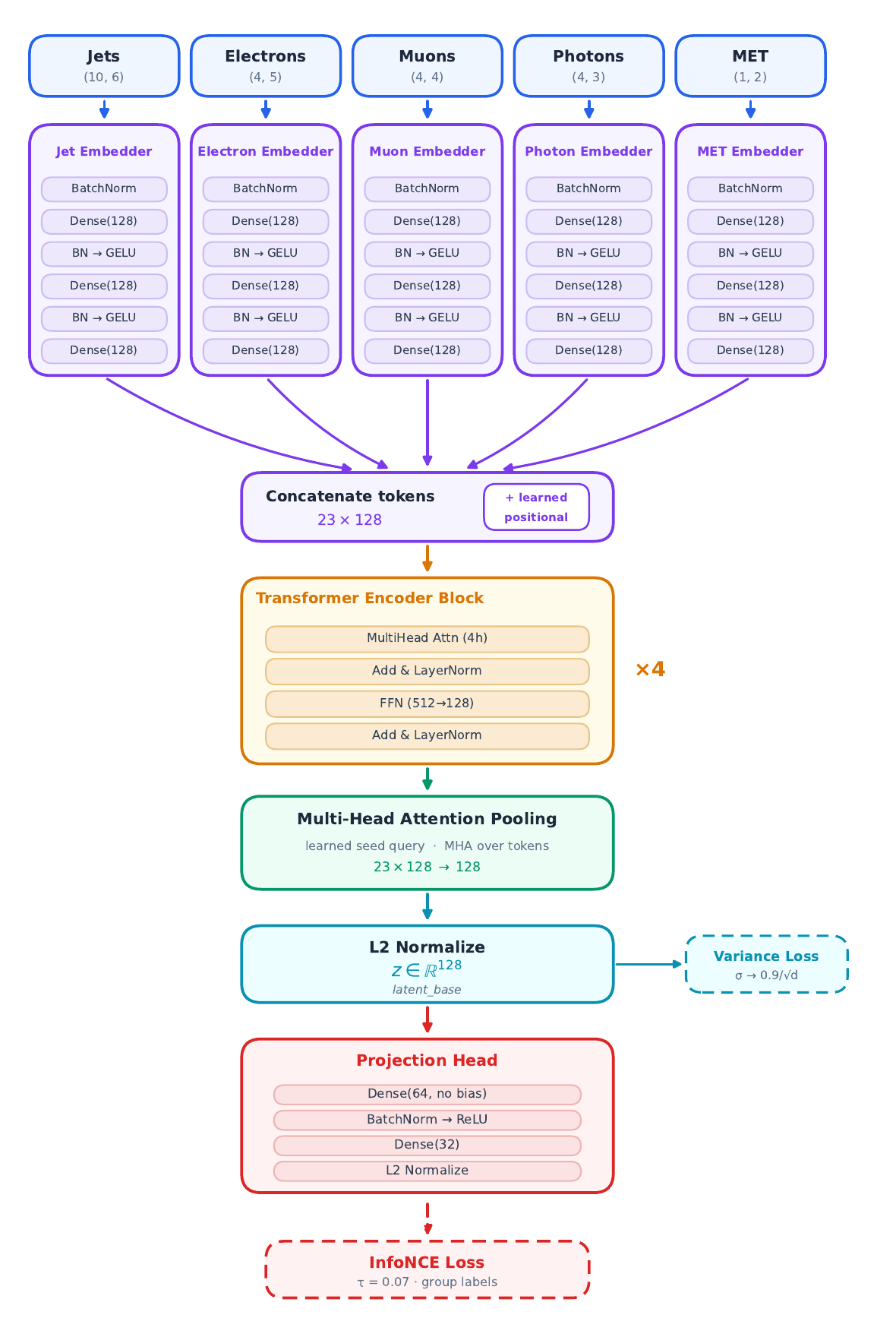}
\caption{Architecture of the Stage 1 ORCA embedder model. Each reconstructed object (jets, electrons, muons, photons, and MET) is mapped to a common 128-dimensional token by a per-type embedding network with shared weights across objects of the same type. The 23 tokens, augmented with learned positional embeddings, pass through four Transformer encoder blocks and a multi-head attention-pooling layer that aggregates them into a single event representation $\vec{r}\in\mathbb{R}^{128}$, normalized onto the unit hypersphere. A projection head maps $\vec{r}$ to an 8-dimensional vector on which the supervised contrastive loss ($\tau=0.07$) is evaluated; a variance regularizer is applied to $\vec{r}$. The projection head is used only during training, and downstream anomaly detection operates on $\vec{r}$.}
\label{fig:model_embedder}
\end{figure*}

Figure~\ref{fig:model_embedder} provides a diagram of the embedder.
The embedder is trained over 48 Standard Model processes from the COLLIDE-2V dataset.
To define the positive pairs of the supervised contrastive loss, the processes are merged into 14 groups of physics processes, listed together with their member processes in Table~\ref{tab:process_groups}; two events are treated as a positive pair if and only if they originate from the same group.
Events are split per process into training and validation sets with an 80/20 ratio.
For processes with fewer than 50\,000 available events, where a proportional split would leave insufficient validation statistics, a fixed validation set of 8192 events is set aside instead and the remainder is used for training.
In total, 4.50 million events are used for training and 1.18 million for validation.

\begin{table*}[tb]
  \caption{The 48 Standard Model processes used to train the embedder (right), merged into the 14 groups that define the labels of the supervised contrastive loss (left). %Group tags in parentheses match the labels used in the figures.
  }
  \label{tab:process_groups}
  \begin{ruledtabular}
  \begin{tabular}{ll}
    Group & Processes \\
    \hline
    \multirow{2}{*}{Single vector boson ($V$)} & $Z\to\nu\bar{\nu}$, $Z\to q\bar{q}$ ($q=u,d,s$), $Z\to b\bar{b}$, $Z\to c\bar{c}$, \\
                                               & $W\to\ell\bar{\nu}$, $W\to q\bar{q'}$ ($q,q'=u,d,c,s$)\\
    Diboson ($VV$) & $WW$, $WZ$, $ZZ$ (all-leptonic, semi-leptonic, all-hadronic) \\
    Triboson ($VVV$) & $VVV$ (all-leptonic, semi-leptonic, all-hadronic) \\
    Drell--Yan (DY) & $Z/\gamma^*\to\ell\ell$ \\
    Low-mass resonance ($\Upsilon$) & $\Upsilon\to\ell\ell$ \\
    Multijet & QCD multijet, $H_\mathrm{T} > 50$~GeV \\
    Prompt photon ($\gamma$) & $\gamma$ \\
    Photon + vector boson ($\gamma V$) & $\gamma + V$ (final state inclusive) \\
    Top pair ($t\bar{t}$) & $t\bar{t}$ (all-leptonic, semi-leptonic, all-hadronic) \\
    \multirow{2}{*}{Gluon--gluon fusion Higgs (ggF $H$)} & $H\to b\bar{b},\, H\to c\bar{c},\, H\to \gamma\gamma,\, H\to gg,\, H\to \tau\tau$, \\ & $H\to WW$ (inclusive),  $H\to ZZ$ (inclusive) \\
    \multirow{2}{*}{Vector Boson Fusion Higgs (VBF $H$)} & $H\to b\bar{b},\, H\to c\bar{c},\, H\to \gamma\gamma,\, H\to gg,\, H\to \tau\tau\,$, \\ &  $H\to WW$ (inclusive),  $H\to ZZ$ (inclusive) \\
    Higgs-strahlung ($VH$) & $WH,\, ZH$ (inclusive) \\
    Di-Higgs ($HH$) & $HH\to 4b,\, b\bar{b}\tau\tau,\, b\bar{b}WW,\, b\bar{b}ZZ,\, b\bar{b}\gamma\gamma$ \\
    Top associated ($t\bar{t}X$) & $t\bar{t}H$, $t\bar{t}t\bar{t}$, $t\bar{t}W$, $t\bar{t}Z$ (final state inclusive) \\
  \end{tabular}
  \end{ruledtabular}
\end{table*}

\begin{figure*}[!tbhp]
\centering
\includegraphics[width=1.3\columnwidth]{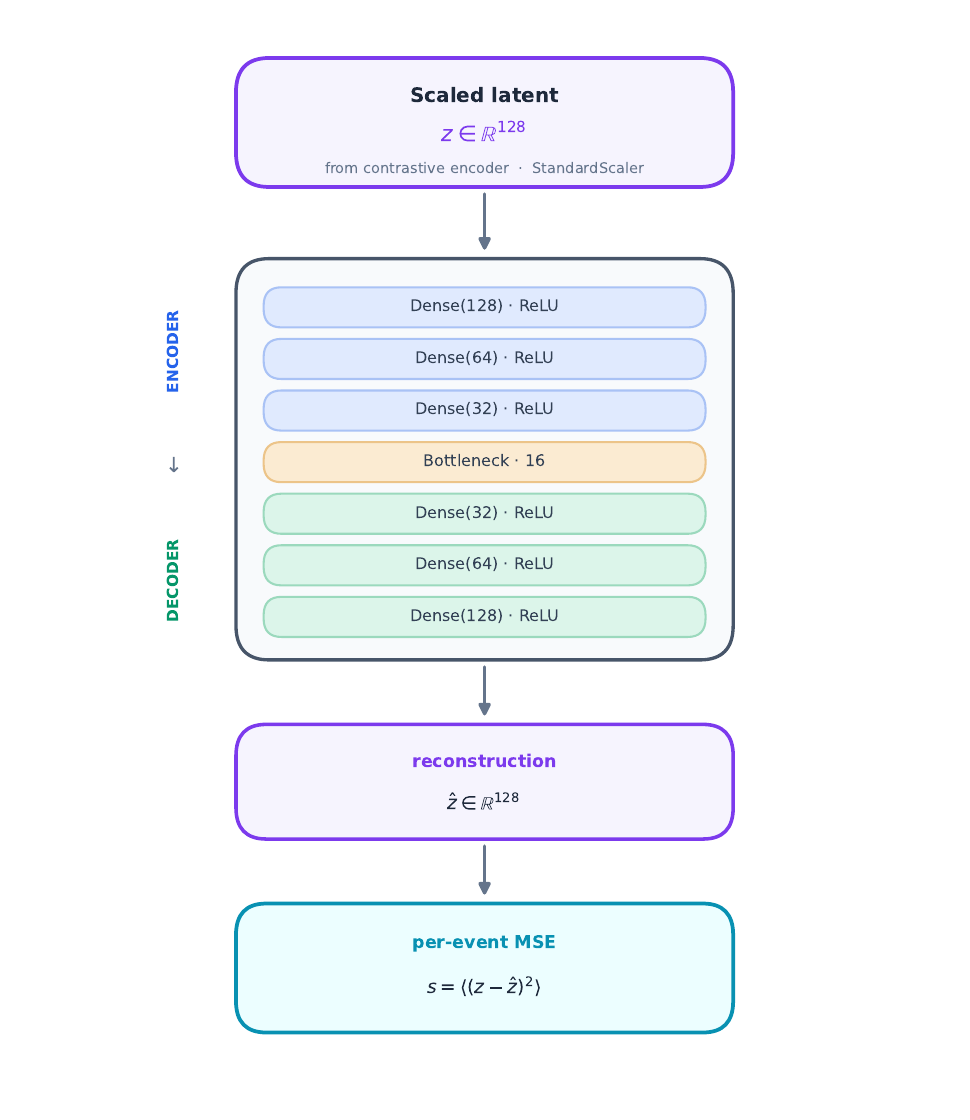}
\caption{Architecture of the Stage 2 ORCA autoencoder model. The 128-dimensional latent representation $z$ produced by the contrastive
encoder is standardized and passed through an autoencoder with encoder layers
of sizes 128, 64, and 32, followed by a 16-dimensional bottleneck. A symmetric
decoder reconstructs the input representation, $\hat{z}\in\mathbb{R}^{128}$.
The per-event anomaly score is defined as the mean squared reconstruction
error, $s=\langle (z-\hat{z})^2 \rangle$.}
\label{fig:model_AD}
\end{figure*}

The model is trained with the AdamW optimizer~\cite{loshchilov2019decoupledweightdecayregularization}, using a weight decay of $10^{-4}$ and global gradient-norm clipping at $1.0$.
The learning rate follows a cosine warm-up schedule: it increases linearly to a peak of $10^{-4}$ over the first $25$ epochs and is then annealed toward zero following a cosine profile over the subsequent $75$ epochs.
Batches are constructed to be group-balanced: each of the 48 training processes contributes a fixed $2048$ events per batch, for a global batch size of $98\,304$ events, which are pooled and shuffled together, so that every physics group is represented in every batch and same-group positives are always available to the supervised contrastive objective.
A dropout rate of $0.5$ is applied throughout the encoder for regularization.
We train for up to $150$ epochs with early stopping, monitoring on the validation loss (patience of $15$ epochs).
The model with the lowest validation loss is restored at the end of training.
%---------------
\paragraph{ORCA Stage 2: Anomaly Detection}
% \tcr{any unsupervised anomaly detection model, such as a density estimator, a normalizing flow, or a distance-based discriminator, could be trained on the same embedding without modifying the first stage, allowing the second stage to be matched to the needs of a given search.
% Here, }
In the second stage, an autoencoder is trained on the embeddings of the first stage to assign an anomaly score to each event.
The projection head is discarded, and each event is represented by its $\ell_2$-normalized 128-dimensional embedding.
The embeddings are standardized per dimension using the mean and standard deviation of the background (e.g. minimum-bias data) sample, which is split into training, validation, and test sets with a 70/15/15 ratio.
The autoencoder architecture (Figure~\ref{fig:model_AD}) is a fully connected network with encoder layers of 128, 64, and 32 nodes with ReLU activations, a linear bottleneck of 16 dimensions, and a mirrored decoder with a linear output layer; all weights carry an $\ell_2$ regularization of $10^{-4}$.

The autoencoder is trained exclusively on background embeddings to minimize the mean squared reconstruction error, using the Adam optimizer with an initial learning rate of $10^{-3}$ and a batch size of 4096.
The learning rate is halved if the validation loss does not improve for 5 epochs, and training is stopped early after 15 epochs without improvement, retaining the best checkpoint.
The anomaly score of an event is defined as the mean squared error between its standardized embedding and its reconstruction,
\begin{equation}
    s(x) = \frac{1}{d} \sum_{i=1}^{d} \left( z_i - \hat{z}_i \right)^2 ,
\end{equation}
where $z$ and $\hat{z}$ denote the embedding and its reconstruction, respectively.
Since the autoencoder learns to reconstruct only the background population of the embedding space, events from processes unlike the background yield large reconstruction errors.

%---------------
\paragraph{Standard Autoencoder Baseline}
To isolate the contribution of the contrastive representation, the ORCA framework is compared against a standard autoencoder trained directly on the input features.
The baseline receives the same standardized 110-dimensional feature vector that is provided to the embedder, consisting of the kinematic features of up to 10 jets, 4 electrons, 4 muons, and 4 photons, together with the missing transverse energy.
Its architecture and training procedure are identical to those of the ORCA anomaly detection stage described above, up to the input dimension: encoder layers of 128, 64, and 32 nodes with ReLU activations, a linear 16-dimensional bottleneck, a mirrored decoder with a linear output layer, and $\ell_2$ weight regularization of $10^{-4}$.
The baseline is trained on the same minimum-bias background sample, split into training, validation, and test sets with the same 70/15/15 ratio, using the same optimizer settings, batch size, learning-rate reduction, and early-stopping criteria, and its anomaly score is likewise the per-event mean squared reconstruction error.
Any difference in performance between the two approaches can therefore be attributed to the representation on which the autoencoder operates: the raw input features for the baseline, and the contrastive embedding for ORCA.

%% file: sections/results.tex
\section{Results}
\label{sec:results}

\begin{figure*}[tb]
\centering
\includegraphics[width=\textwidth]{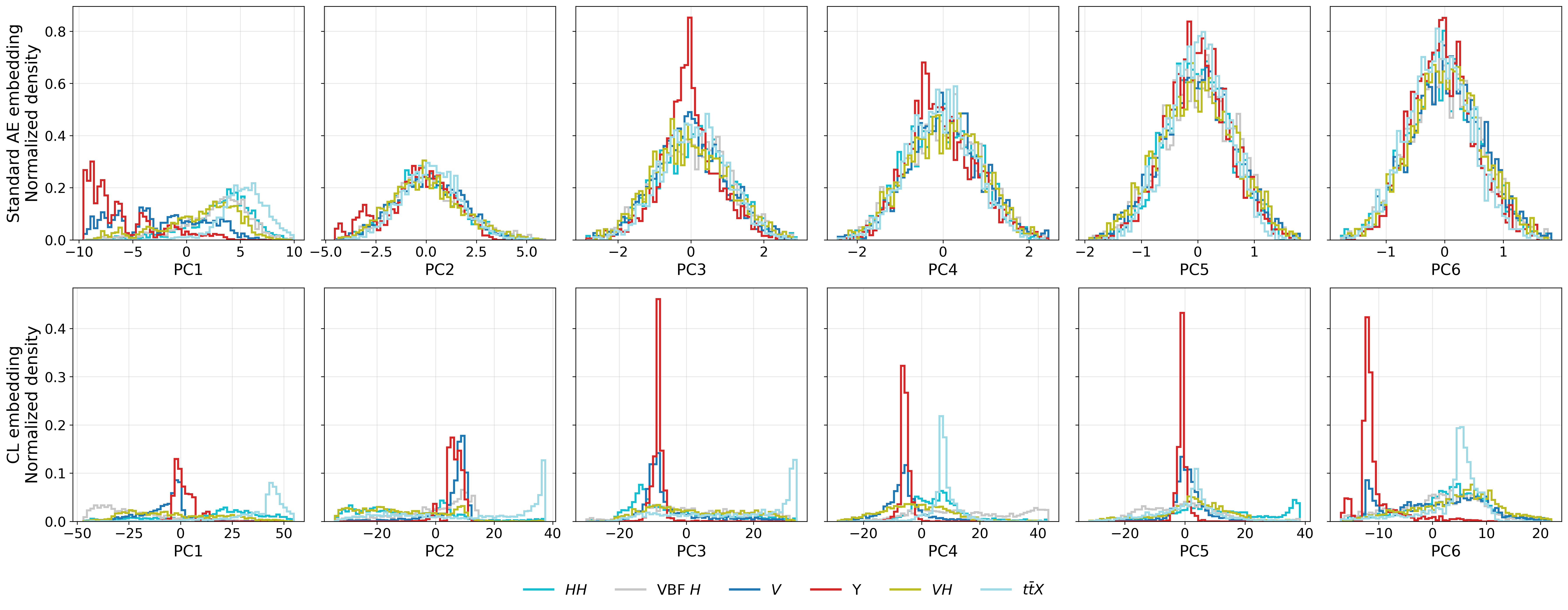}
\caption{Primary component analysis in six dimensions of the standard autoencoder (top) and ORCA Stage 1 embedding model output space (bottom). Only a focused subset of all training signal models is shown. The capacity of the contrastive learning in the ORCA approach allows for more distinguished classes of signals in the latent space, ultimately providing more information to a downstream autoencoder training.
\label{fig:latentspace} }
\end{figure*}

As the introduction of a supervised contrastive embedder to the standard autoencoder-based AD methodology is chosen to better structure the latent space for the AD task, an examination of the latent space representations of various SM processes provides a key view into what the model learned.
Figure~\ref{fig:latentspace} provides a visualization of the ORCA embedding space via a primary component analysis (PCA) in six dimensions.
Only a representative set of training signal models are shown for visual clarity.
The ORCA embedding space, provided by Stage 1 training, is shown in comparison to the latent space of the standard autoencoder.
ORCA demonstrates a considerably enhanced ability to separate different classes of signals along the PCA components, with several distinct peaks visible compared to the broad overlapping distributions of many signals in the autoencoder latent space.
% \tcr{Note differences in the two approaches for DY/upsilon?}

The clustering of information in the latent space provided by the contrastive learning in ORCA is essential, as it can be used for the task of anti-background selection, which in practice drives AD performance.
Further, it enables the application of standard interpretation techniques in HEP directly to the embedding space, allowing for exploitation of high-dimensional information in the interpretation process.
This allows for an AD method whose final output score is explainable, namely by comparing a reserved test set to the training signal classes in the embedding space.
In this way ORCA can be used to classify a test set based on its proximity to other signal classes, offering a window into understanding potential anomalous findings in data.
Details of these two claims are given in the following sections.

%-------------------------------------------------
%-------------------------------------------------
\subsection{Anomaly Detection Performance}

Performance of an anomaly detection method can be assessed by the breadth of process sensitivity provided by thresholding on the model's output loss as evaluated on a test set.
In this case, the set of signal models used in training are considered to be representative of the processes accessible at the LHC.
Two different metrics are considered.
First, the area under curve (AUC) of the receiver operating characteristic (ROC) represents the model's general ability to distinguish a given signal from minimum-bias background data.
Second, the signal efficiency (aka TPR, the true positive rate) at a low fixed background acceptance (aka FPR, the false positive rate), in this case 10$^{-4}$, is used to study the model's performance in the high background rejection regime where most physics analyses must be done to suppress the huge background rates.

\begin{figure}[!tb]
\centering
\includegraphics[width=\columnwidth]{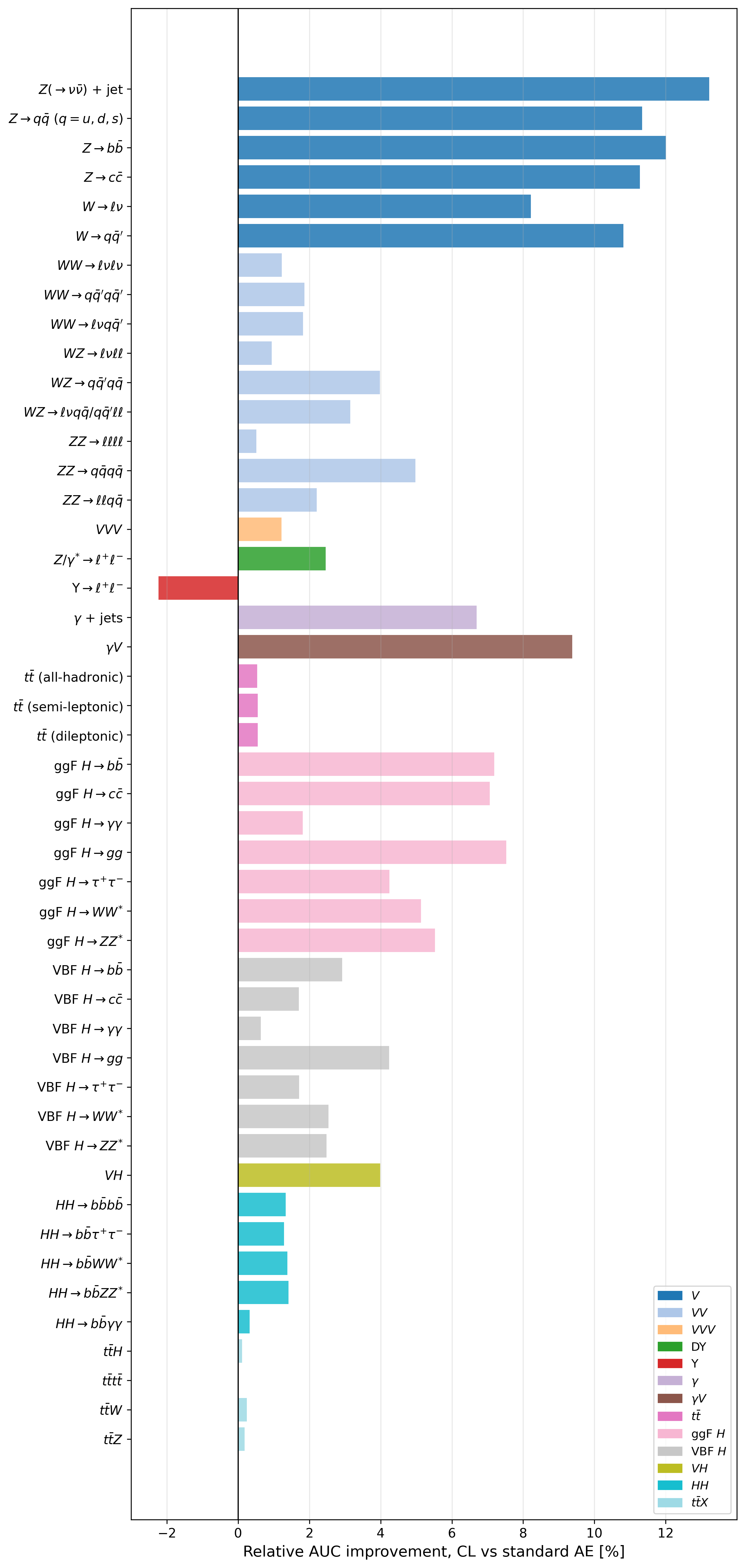}
\caption{The relative percent improvement of the contrastive method with respect to the standard autoencoder in terms of AUC. All groups defined in Table~\ref{tab:process_groups} are shown except QCD due to its proximity to the background.
\label{fig:perf_aevscl_auc} }
\end{figure}

\begin{figure}[!tb]
\centering
\includegraphics[width=\columnwidth]{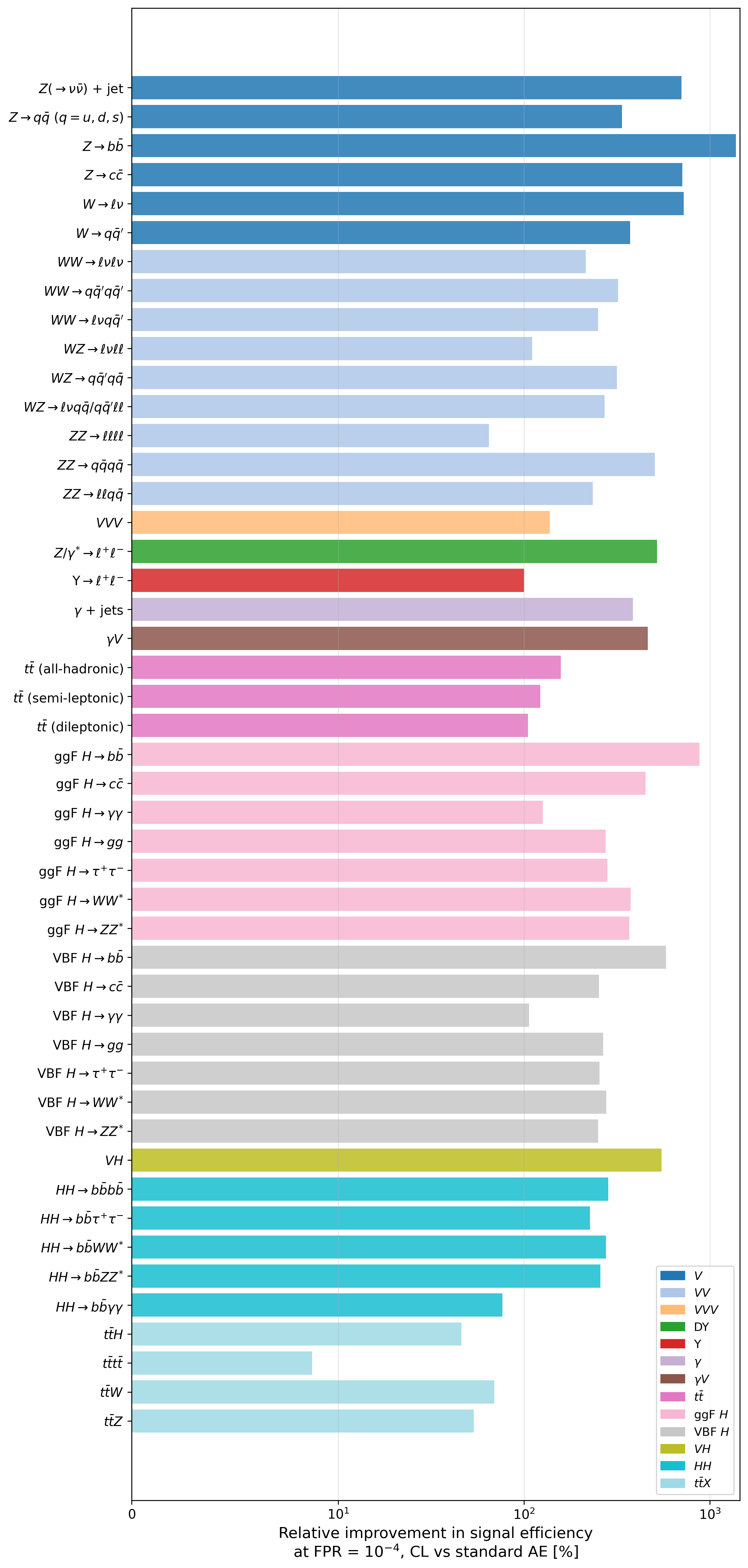}
\caption{The relative percent improvement of the contrastive method with respect to the standard autoencoder in terms of TPR at a fixed FPR of 10$^{-4}$. All groups defined in Table~\ref{tab:process_groups} are shown except QCD due to its proximity to the background.
\label{fig:perf_aevscl_tpr} }
\end{figure}

\begin{figure*}[!tb]
\centering
\includegraphics[width=0.47\textwidth]{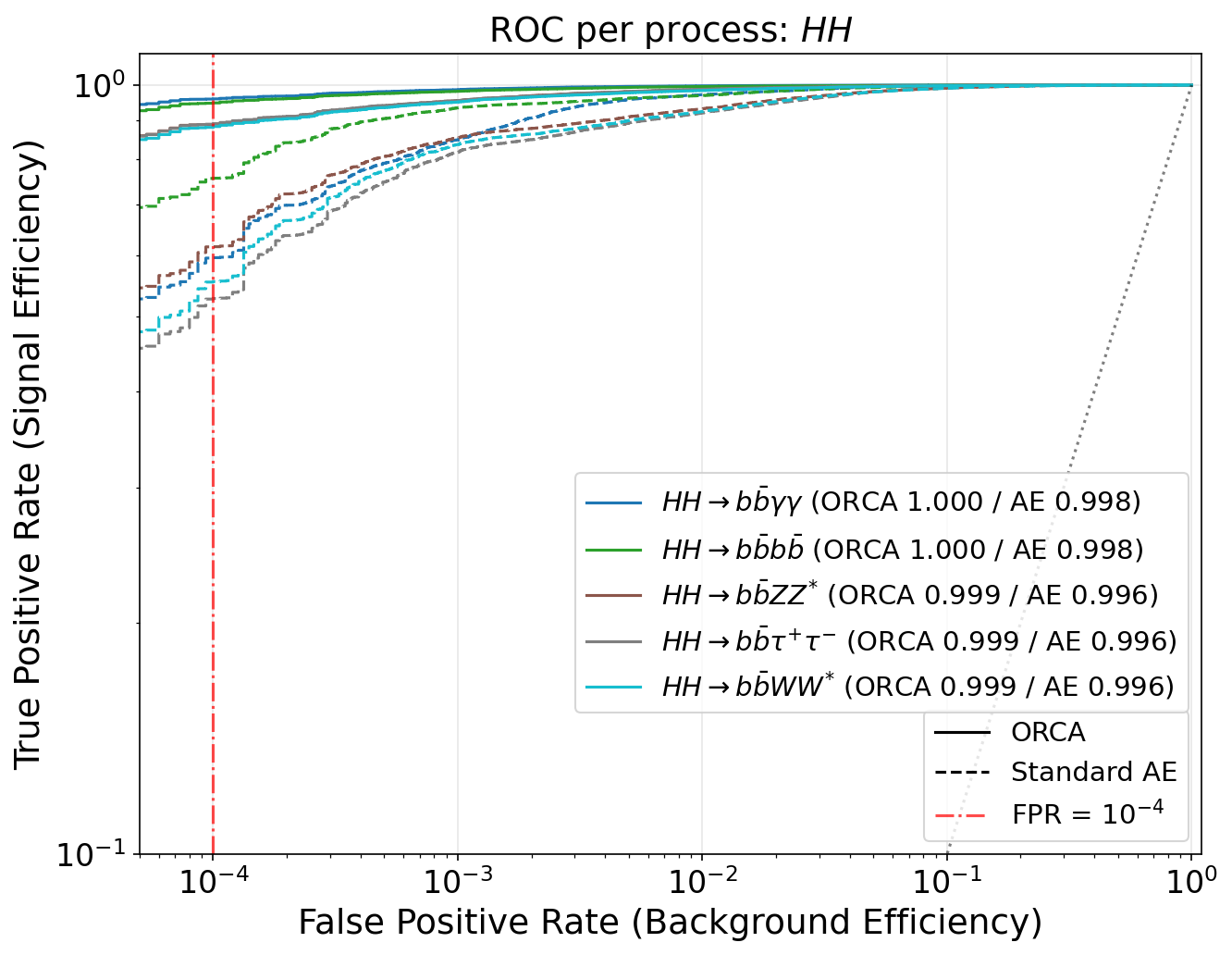}
\includegraphics[width=0.47\textwidth]{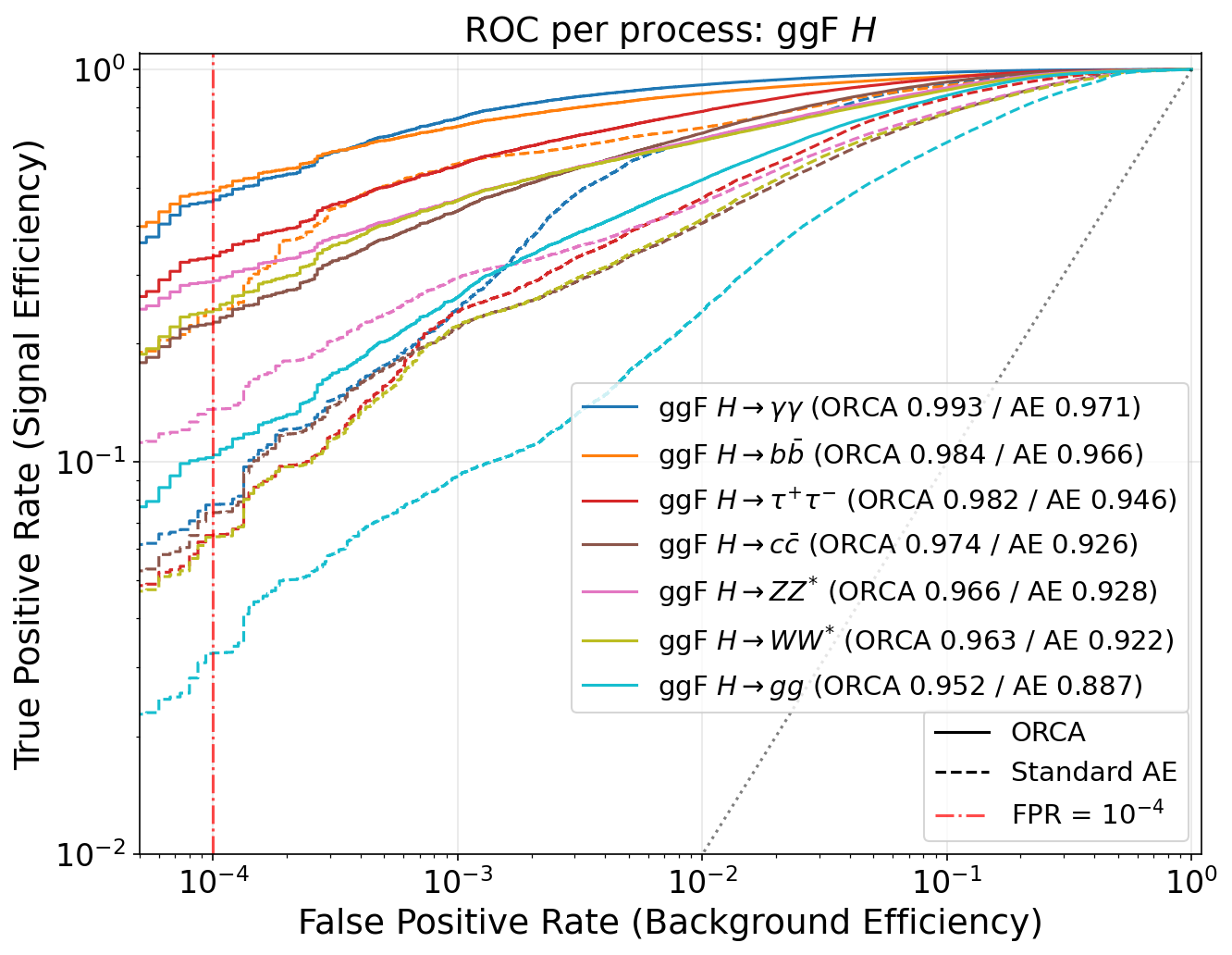}
\includegraphics[width=0.47\textwidth]{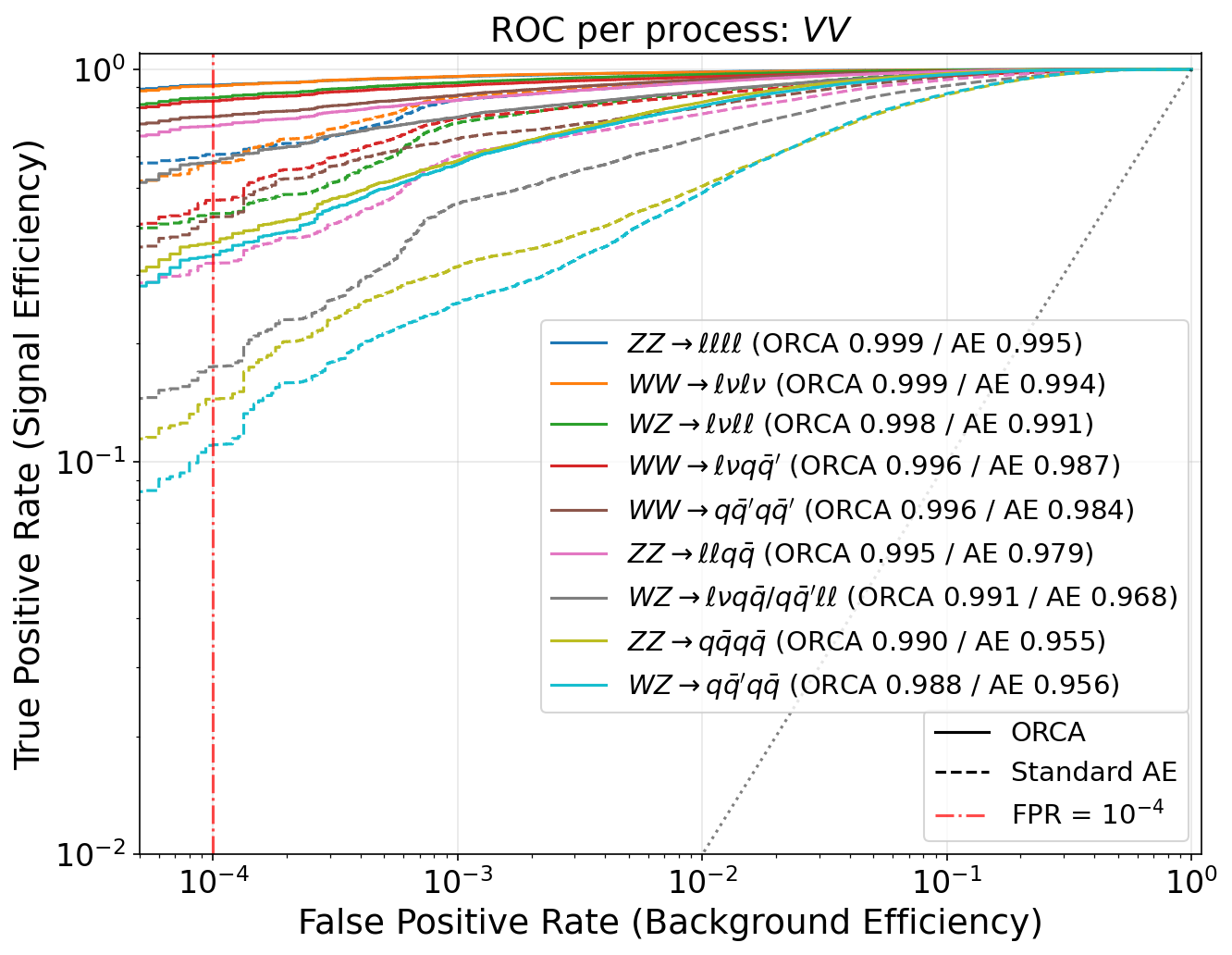}
\includegraphics[width=0.47\textwidth]{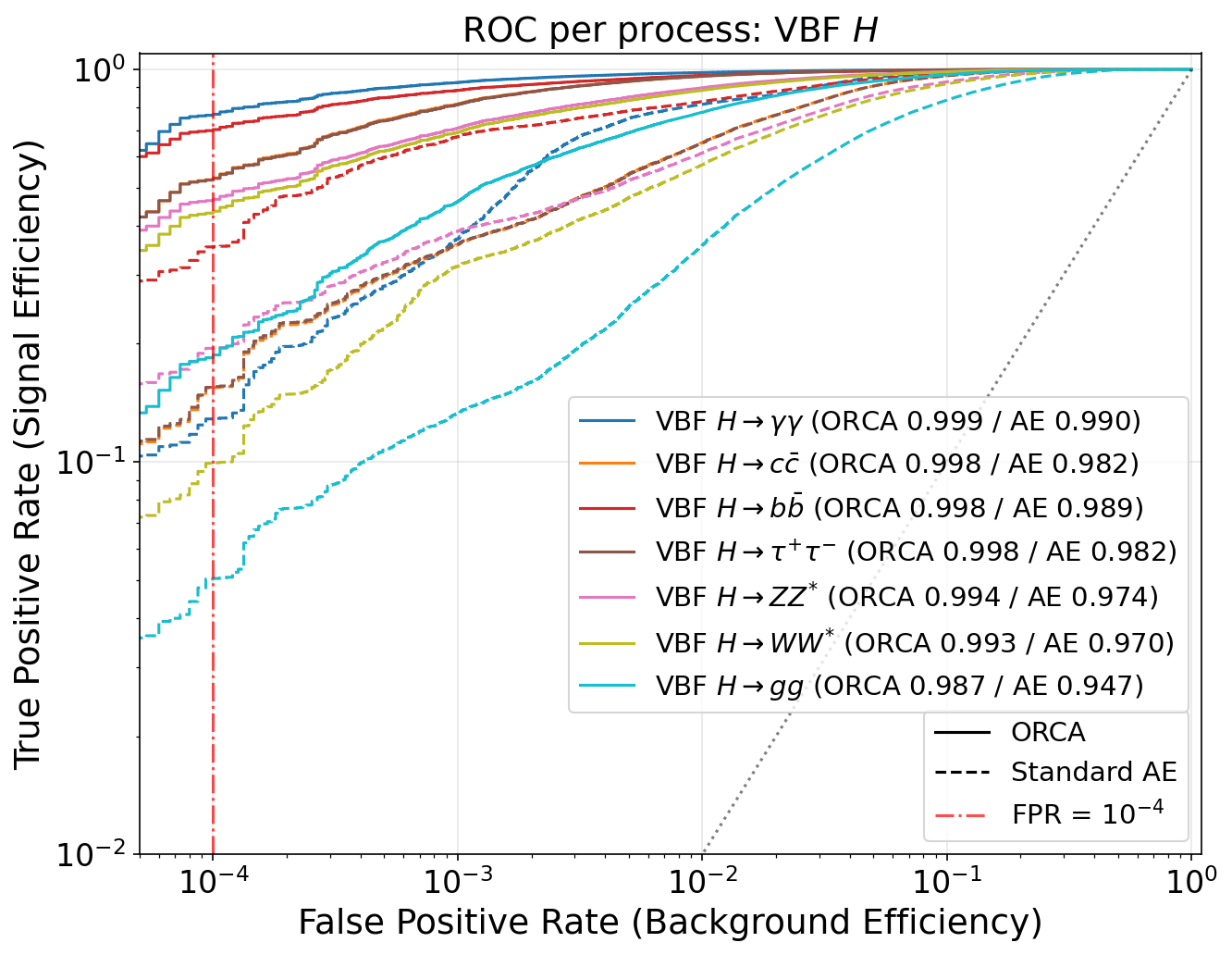}
\caption{ROC curves comparing ORCA (solid) and the standard autoencoder (dash), shown for four representative process groups: Di-Higgs (top left), Gluon-gluon fusion Higgs (top right), Two Vector Boson (bottom left), and Vector Boson Fusion Higgs (bottom right). The signal efficiency is shown as a function of the background efficiency, evaluated against minimum-bias data. The AUC for both models is provided parenthetically in the legend for each process. The dashed vertical line indicates the background efficiency working point of $10^{-4}$ used for the fixed background rejection comparison.
\label{fig:perf_aevscl_roc} }
\end{figure*}

Figure~\ref{fig:perf_aevscl_auc} and Figure~\ref{fig:perf_aevscl_tpr} shows the percent gain in AUC and fixed background rejection TPR respectively for all signal models used in training.
Figure~\ref{fig:perf_aevscl_roc} shows the full ROC curves for four representative signal processes, illustrating the performance difference across the complete range of background acceptance.
ORCA is compared with respect to the standard autoencoder, isolating the impact of the contrastive learning-informed embedding space on AD performance.
Significant gains are noted across nearly all signals in both metrics.
The sole exception is noted for the $\Upsilon(\ell\ell)$ process in AUC, which the ORCA embedder places close to minimum-bias data in the embedding space, thus ranking it more background-like in the bulk phase space.
However, considering the high background rejection regime provided by the TPR metric, the contrastive method recovers a strong performance gain over the autoencoder.

%-------------------------------------------------
%-------------------------------------------------
\subsection{Interpretability}

The main motivation for using contrastive learning in the embedder is that it enables a quantitative interpretation of the selected anomalous events, in the form of a per-process decomposition of the selected sample obtained from a template fit in the embedding space.
The supervised contrastive objective encourages the known processes to occupy different regions of the embedding space, so that their distributions, while still partially overlapping, are sufficiently distinct for the fit to constrain the composition of a selected sample.
In an embedding without this objective, such as one trained purely for reconstruction, the process distributions overlap strongly and the corresponding templates are nearly degenerate, limiting the ability of the fit to attribute the observed events to individual processes.

The interpretation method proceeds as follows.
We first reduce the dimensionality of the embedding space with a principal component analysis (PCA), retaining the leading components up to a fixed variance threshold, and then apply independent component analysis (ICA) to the retained components, which reduces the statistical dependence between the resulting dimensions.
For each known process, each retained dimension $d$ is histogrammed using labeled simulated samples, and the normalized histograms define one probability mass function (PMF) per dimension and per process.
We model the expected event count in bin $b$ of dimension $d$ as a mixture,
\begin{equation}
\mu_{d,b}(\theta) = \nu_{\mathrm{bkg}} \, \hat{f}^{(\mathrm{bkg})}_{d,b}
+ \sum_{i=1}^{S} \nu_i \, \hat{f}^{(\mathrm{sig}_i)}_{d,b},
\end{equation}
where $\hat{f}^{(\mathrm{bkg})}_{d,b}$ and $\hat{f}^{(\mathrm{sig}_i)}_{d,b}$ are the template PMFs, $S$ is the number of signal processes, and the parameters of interest $\theta = (\nu_{\mathrm{bkg}}, \nu_1, \ldots, \nu_S)$ are the background and signal yields determined by the fit.
Each bin count $n_{d,b}$ is treated as an independent Poisson random variable, and the pseudo-likelihood is the product of the per-bin Poisson terms over all dimensions and bins,
\begin{equation}
\mathcal{L}(\theta)=\prod_{d=1}^{D}\prod_{b=1}^{B}
\frac{\mu_{d,b}(\theta)^{\,n_{d,b}}\,e^{-\mu_{d,b}(\theta)}}{n_{d,b}!},
\end{equation}
where $n_{d,b}$ is the observed number of events in bin $b$ of dimension $d$, $D$ is the number of retained embedding dimensions, and $B$ is the number of bins per dimension.
Because each event contributes to one bin in every dimension, the $D$ per-dimension factors are not statistically independent, and residual dependence between embedding dimensions is neglected by construction, so $\mathcal{L}(\theta)$ is a composite likelihood rather than a genuine one.
The yields are estimated by minimizing the negative log-likelihood (NLL),
\begin{equation}
-\ln\mathcal{L}(\theta)=\sum_{d=1}^{D}\sum_{b=1}^{B}
\left[\mu_{d,b}(\theta)-n_{d,b}\ln\mu_{d,b}(\theta)\right]+\mathrm{const},
\end{equation}
subject to $\nu_{\mathrm{bkg}}\geq 0$ and $\nu_i\geq 0$.
Since each event enters $D$ factors, the factorized product overstates the information content of the sample, and intervals derived from the inverse Hessian of the NLL would generally undercover.
The statistical uncertainties on the fitted yields are therefore obtained from the robust (sandwich) covariance estimator for composite likelihoods~\cite{Varin:2011:CompositeLikelihood}:
\begin{subequations}
\begin{align}
\widehat{\mathrm{Var}}(\hat\theta) &= H^{-1}\,\hat{J}\,H^{-1},
\\
\hat{J} &= \sum_{e=1}^{N} s_e\, s_e^{\mathsf{T}},
\\
s_e^{(k)} &= \sum_{d=1}^{D}
\frac{\hat{f}^{(k)}_{d,\,b_d(e)}}{\mu_{d,\,b_d(e)}(\hat\theta)},
\end{align}
\end{subequations}
where $H$ is the Hessian of the NLL at the minimum, $\hat{J}$ is the empirical covariance matrix, $b_d(e)$ is the bin of event $e$ in dimension $d$, and the index $k$ runs over the fit parameters, with $\hat{f}^{(k)}$ denoting the background template for $k = \nu_{\mathrm{bkg}}$ and the template of signal process $i$ for $k = \nu_i$.
The sum in $\hat{J}$ is left uncentered so that the Poisson fluctuation of the total yield is included.
The resulting intervals are asymptotic and are not quoted for yields at the physical boundary $\nu_i = 0$.
The statistical uncertainty of the template PMFs and the dependence on the ICA definition are not propagated.

More fundamentally, the template is a limited set of simulated processes and cannot span the full space of possible new-physics signals.
The fitted yields should therefore be read as a projection of the anomalous data onto the available reference processes: a signal absent from the library is absorbed by the templates it most resembles, so the yields characterize the data rather than identify its origin.
The method thus serves as a diagnostic of the composition of an anomalous region, and the signal-injection studies quantify its accuracy for signals that are represented in the signal set.

The fit procedure is validated on pseudo-data constructed from simulated events, using subsets that are statistically disjoint from those used to build the template PMFs.
In a realistic application, the templates would still be built from simulated background and signal samples, while the fitted sample would consist of recorded event data, so the accuracy of the extracted yields would depend on how well the simulation describes the data. Since the pseudo-data used here derive from the same simulation as the templates, these tests probe the statistical performance of the method alone.
To demonstrate the ability of ORCA to correctly attribute events from a known source, the test set of each signal model used in training is provided as pseudo-data to the fitting procedure.

%Two different types of fit tests are performed: informed, where the signal template of the test set is used in the fit, and blind, where the correct signal template is not provided.
%These two sets are designed to show that ORCA can correctly determine the true source of a test set when provided the correct template, and further to understand what features make signals proximal to one another in the embedding space.
To assess how the template fit behaves for signals that are and are not represented in the template library, we first perform a holdout test.
In each test, one signal group is excluded from the contrastive training, so the embedding has never seen it, and 200 of its events passing the AD threshold (corresponding to an FPR of $10^{-3}$) are injected into the anomaly-passing background sample.
The fitted dataset therefore contains $N_{\mathrm{bkg}} + 200$ events, where $N_{\mathrm{bkg}}$ is the number of background events passing the AD selection and varies between holdout tests, since each test employs its own embedder and autoencoder trained without the holdout group.
The injected sample is then fitted twice: a \emph{blind} fit, in which the template set does not include the injected group, and an \emph{informed} fit, which additionally includes its template, built from the remaining holdout events.
Because the fit maximizes an extended likelihood in which all template normalizations are free, the fitted yields in either configuration sum to the total number of fitted events.
The blind fit shows which known templates absorb an unrepresented signal, while the informed fit tests whether the yield of a represented signal is recovered.

\begin{figure*}[!tb]
\centering
\includegraphics[width=0.47\textwidth]{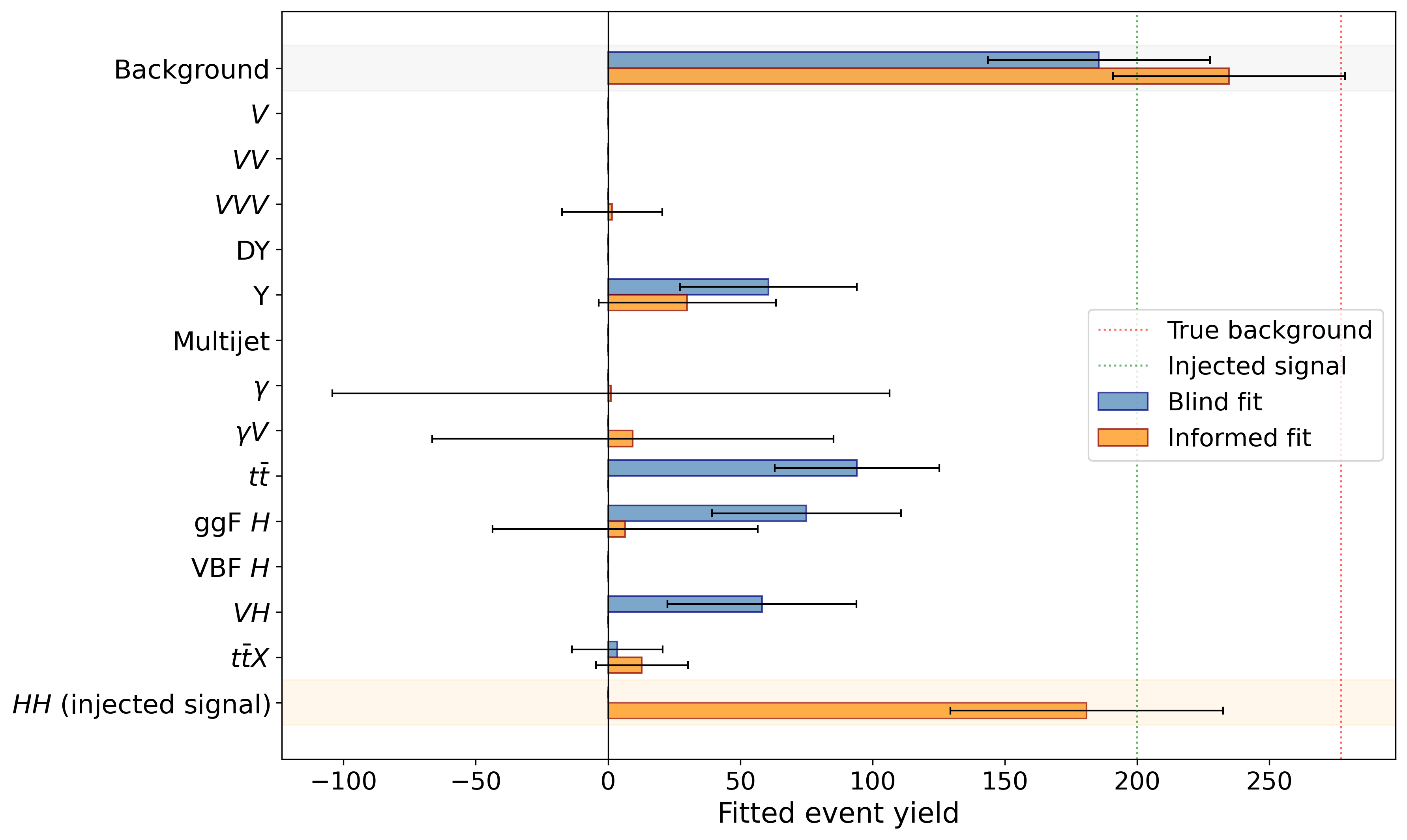}
\includegraphics[width=0.47\textwidth]{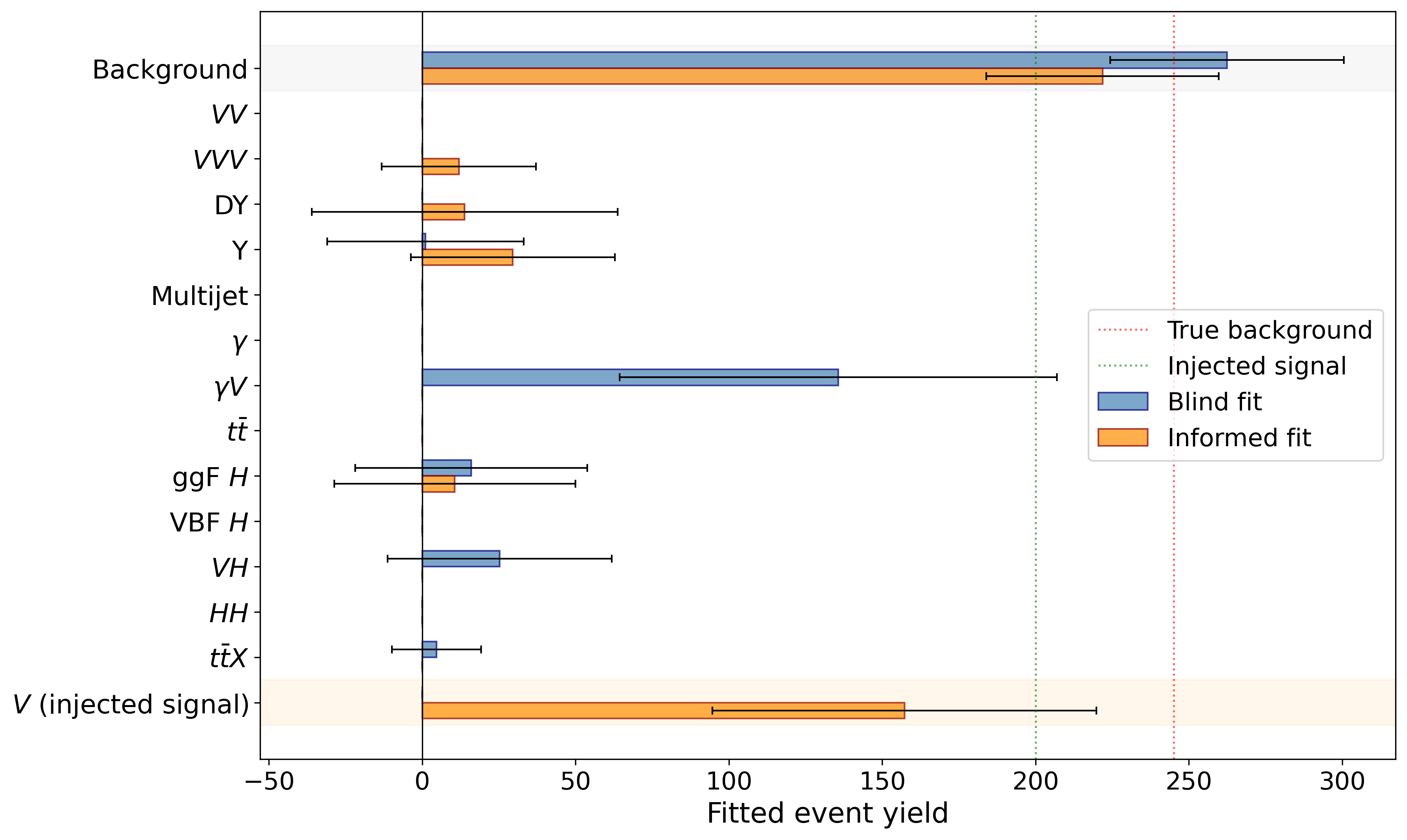}
\includegraphics[width=0.47\textwidth]{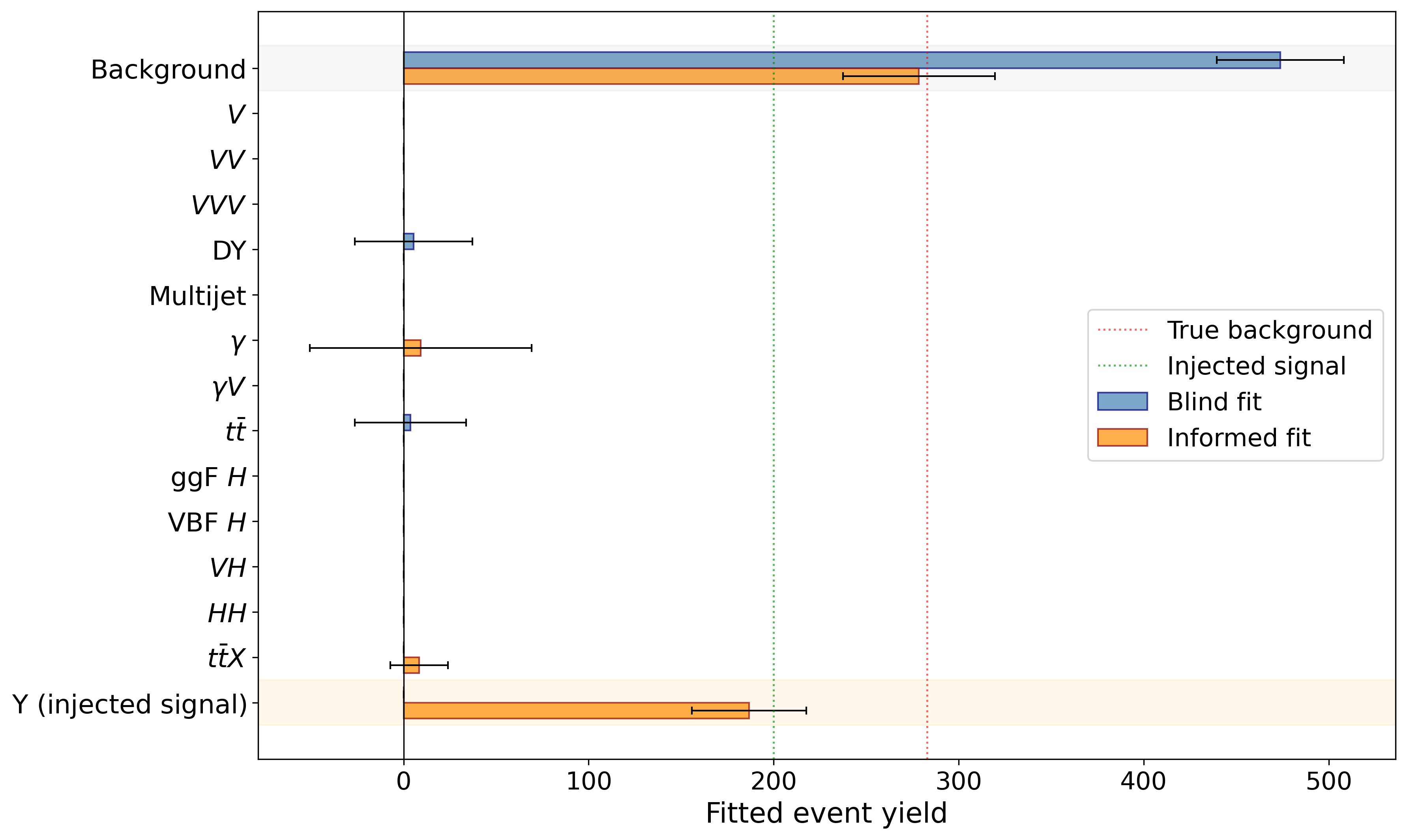}
\includegraphics[width=0.47\textwidth]{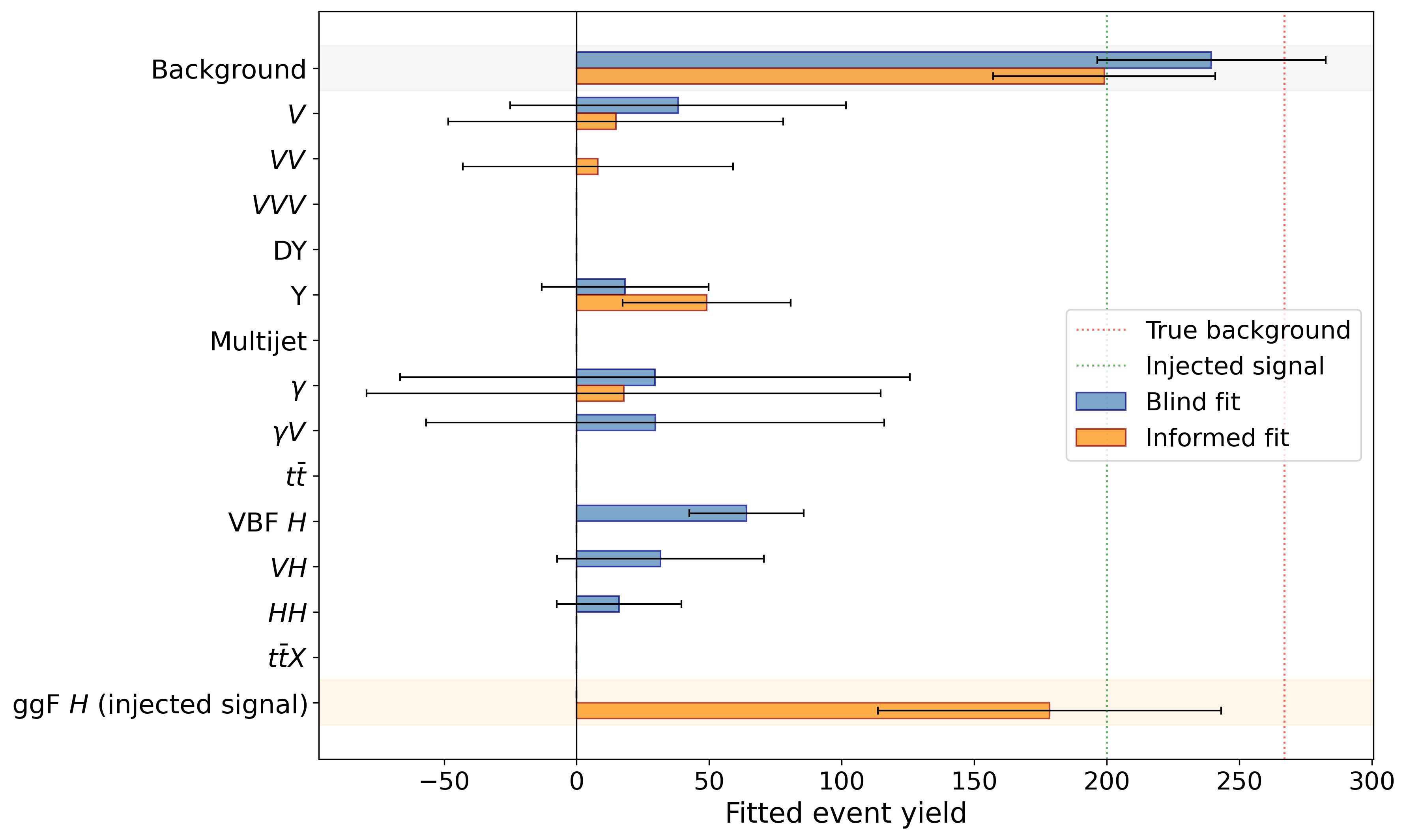}
\caption{Fitted event yields in the holdout test for four groups: Di-Higgs (top left), Single Vector Boson (top right), Low-mass resonance $\Upsilon$ (bottom left), and Gluon–gluon fusion Higgs (bottom right); in each case the group is excluded from training and 200 of its AD-passing events are injected. Within each panel, each row is one fit parameter: the background yield (shaded gray) and the per-group signal yields, with the injected group highlighted. Blue bars show the blind fit (no template for the injected group); orange bars the informed fit (its template included). Error bars are the sandwich estimates; yields at the boundary $\nu_i = 0$ carry no interval. Vertical dotted lines mark the true background yield (red) and the injected yield (green).
\label{fig:fittests}}
\end{figure*}
The procedure is repeated with each signal group as the holdout.
Figure~\ref{fig:fittests} shows the resulting yields for four representative groups from the holdout test.
For each these four groups, in an informed fit, all injected events are correctly attributed to the simulated signal process within error.
The blind fit provides a window into the learned embedding space structure of physics processes.
For example, Di Higgs process events are categorized primarily as top pair in the absence of the correct template, likely revealing that the model learns final state multiplicity in its embedding training.
Single Vector Boson events are similarly well-recovered in the informed fit; in the blind fit they are primarily categorized as $\gamma+V$, differing only from the truth process in the presence of a final state photon, indicating strong learning of physics process characteristics.
Low mass resonance $\Upsilon$ events are nearly fully recovered in an informed fit, while in the blind fit they are absorbed dominantly as the background, indicating ORCA learns low mass resonance in the embedding training.
These results indicate that ORCA can be used for a more sophisticated characterization of a potential excess in an AD signal region; whereas previously excesses could only be described by their features in collider observable space, the ORCA blind test enables categorization into simulated physics processes based on learned correlations of such one-dimensional observables.
Additionally, the informed test results offer a new approach to re-interpretation of an AD search, wherein a new simulated model could be introduced for statistical tests of compatibility with respect to an unblinded data selection.
\begin{figure*}[!tb]
\centering
\includegraphics[width=0.47\textwidth]{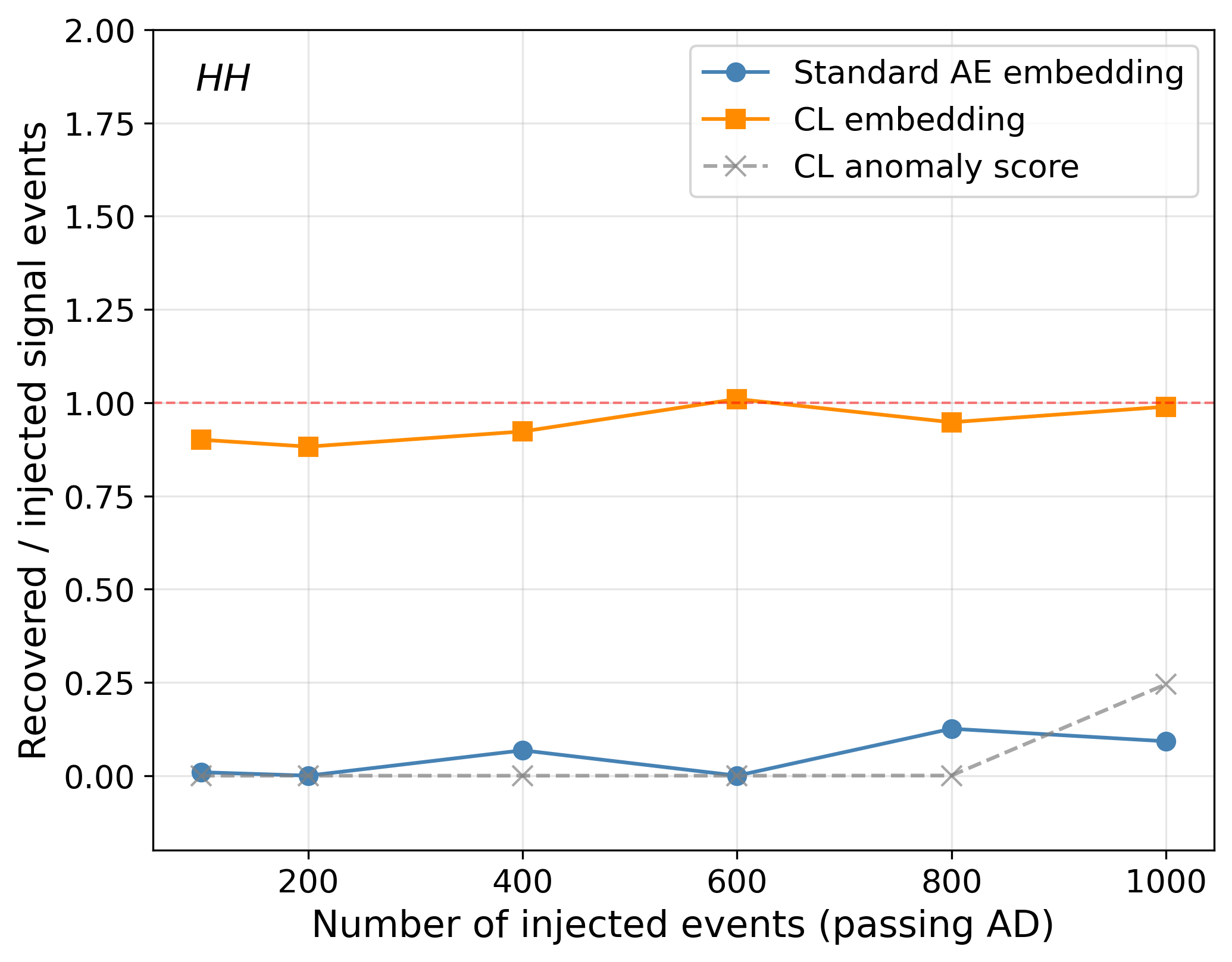}
\includegraphics[width=0.47\textwidth]{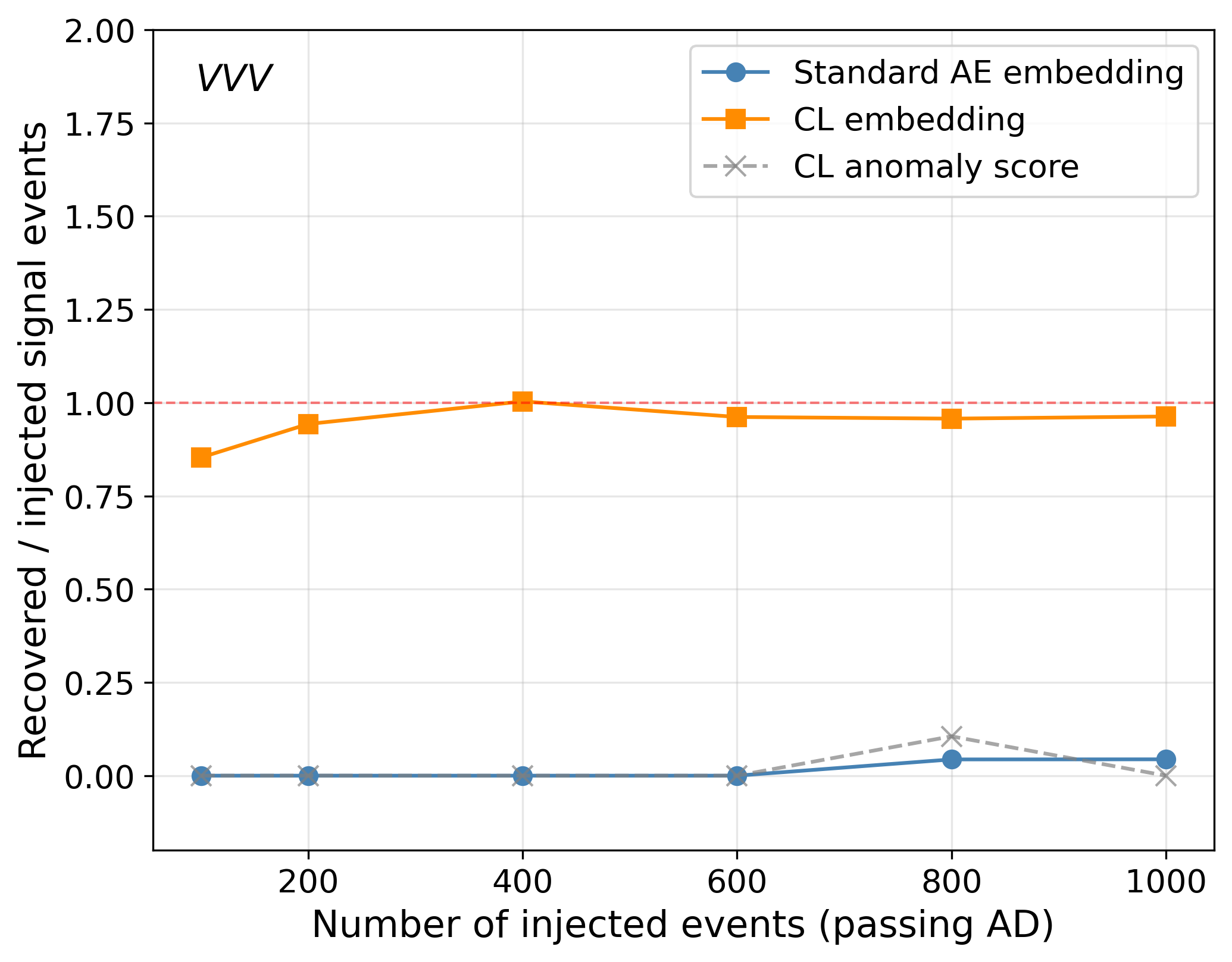}
\includegraphics[width=0.47\textwidth]{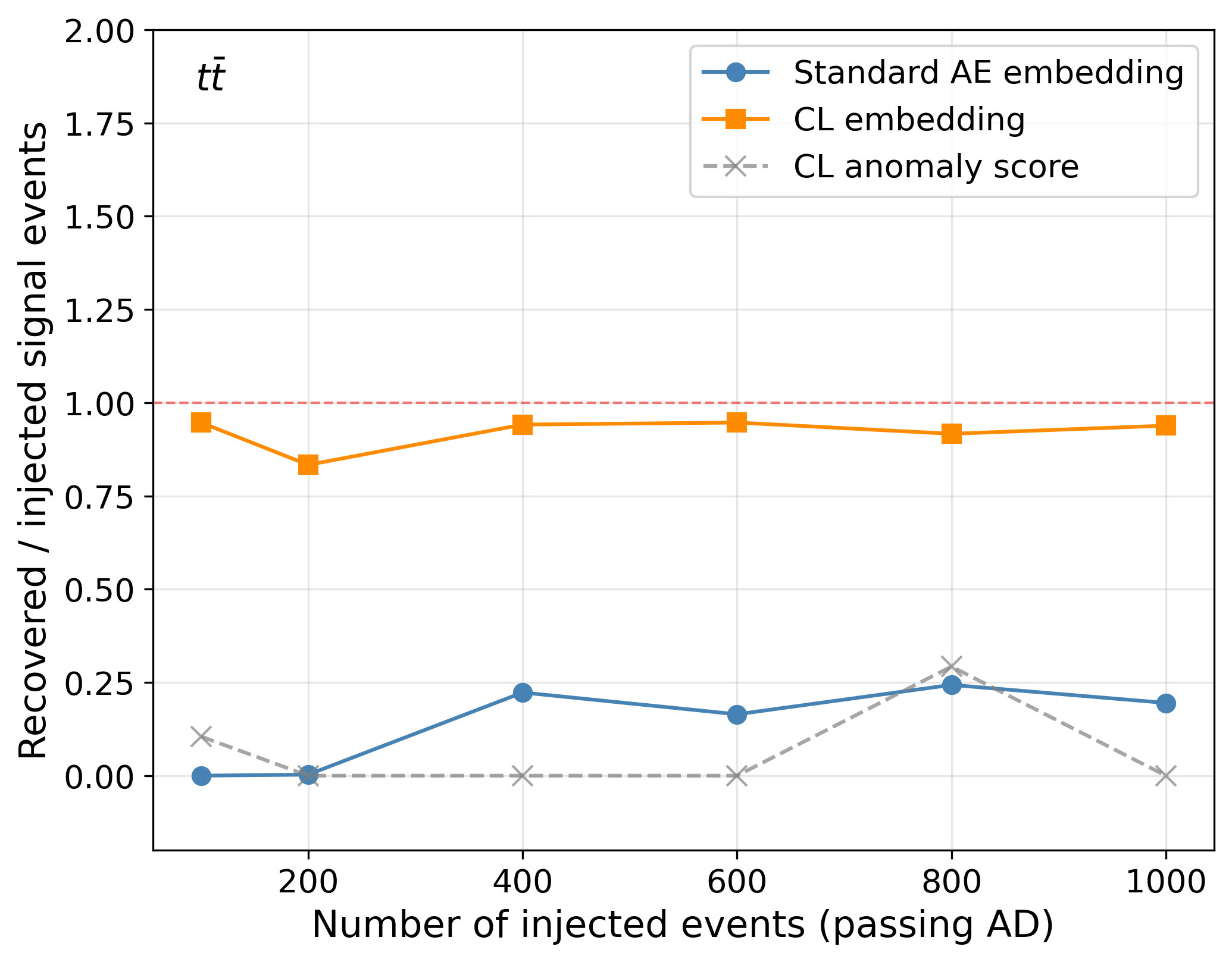}
\includegraphics[width=0.47\textwidth]{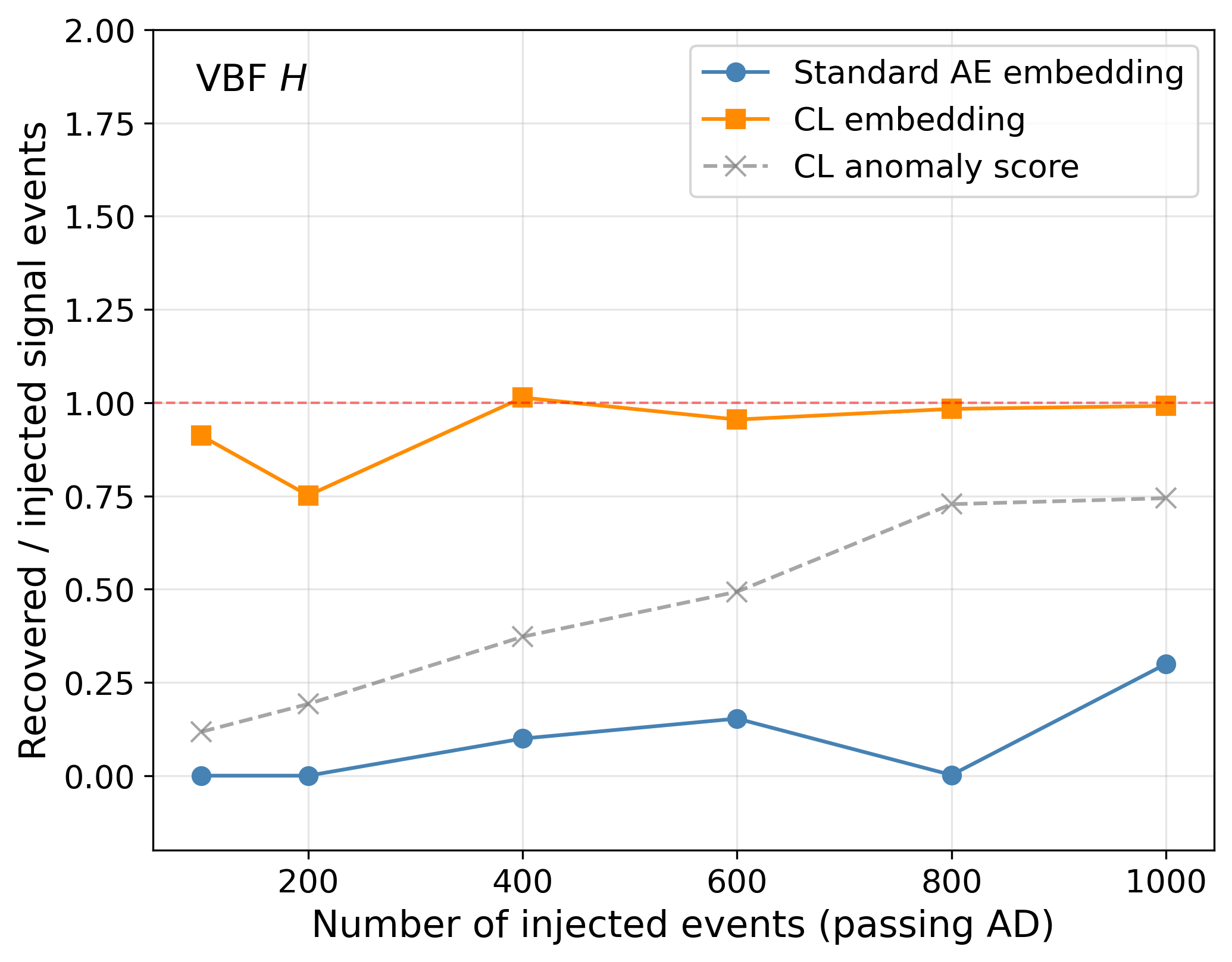}
\caption{Signal recovery, defined as the fitted signal yield divided by the number of injected events, as a function of the number of injected signal events for four representative process groups: Di-Higgs (top left), Three Vector Boson (top right), Top pair (bottom left), and VBF Single Higgs (bottom right). Fits are performed in the standard autoencoder embedding (blue circles), the contrastive learning embedding (orange squares), and on the one-dimensional contrastive learning anomaly score (gray crosses). The dashed red line indicates exact recovery of the injected signal.
\label{fig:signalrecovery} }
\end{figure*}
Next, to quantify the accuracy of the fitted yields for signals that are represented in the template library, we perform signal injection tests.
In each test, a varying number of signal events, drawn from the subset of a group's events passing the AD threshold set at a background efficiency of $10^{-3}$ and ranging from 100 to 1000, is injected into the background sample, and the fit is performed with the corresponding signal template included.
The recovery is defined as the fitted signal yield divided by the number of injected events, so that exact recovery corresponds to unity.
We aim to isolate the effect of two design choices in ORCA: first the use of CL in the embedder, and second the use of the multi-dimensional latent space for likelihood fitting instead of a single output dimension.
In this aim, the signal injection test is repeated in three fitting spaces: the contrastive learning embedding; the one-dimensional contrastive learning anomaly score, to highlight the benefit of fitting a multi-dimensional space; and the latent space of a standard unsupervised autoencoder, to highlight the benefit of the contrastive objective.

Figure~\ref{fig:signalrecovery} shows the recovery as a function of the number of injected events for four representative process groups.
%\tcr{JG: again needs discussion here: what should the reader take away?}
In each case, the CL multi-dimensional embedding space fit provides the highest recovery across injected signal yields.
The nearly negligible signal recovery from the CL anomaly score indicates the power of the interpretation method proposed here compared to standard reinterpretation approaches that rely on fits to one-dimensional observable distributions.
We can further conclude that the use of CL is an essential component of creating such a multi-dimensional latent space with relevant physics categorization; the signal recovery of the CL embedding fit also far exceeds that of the multi-dimensional baseline autoencoder latent space, indicating that the extra dimensions alone are not enough to drive the best performance.

%% file: sections/conclusions.tex
\section{Conclusions}
\label{sec:conclusions}

This work demonstrates the utility of contrastive learning for high-performance and interpretable anomaly detection at high energy colliders. 
The ORCA method leverages a two-stage architecture, first using labeled signal events to create a latent space structured by physics process knowledge, and second using this embedding to create a per-event anomaly score from an autoencoder with a reconstruction objective. 
The final ORCA anomaly score outperforms a baseline autoencoder across AUC and fixed background rejection metrics for a wide variety of Standard Model processes.
Furthermore, this work demonstrates a novel approach to anomaly detection interpretation through the direct likelihood fitting of the ORCA embedding space, made possible by the high performance process clustering provided by CL. 
This approach allows for the categorization of events of an unknown source into template processes, and outperforms standard single-dimensional observable likelihood fitting in signal injection tests. 
Together these capabilities offer a path to interpret data artifacts and re-interpret signal regions with new simulated signal models, substantially enhancing the impact of AD searches at colliders.
Finally, the two-stage structure of ORCA is modular. The transformer-based embedder could be replaced by a lightweight architecture suited to low-latency environments such as the hardware trigger, and the autoencoder by any anomaly detection model operating on the embedding. The framework can therefore be tailored to the constraints of a given deployment while retaining the interpretability.